\documentclass[journal]{IEEEtran}
\usepackage{ifpdf}
\usepackage{cite}
\usepackage{amsmath,amssymb,amsfonts}
\usepackage{textcomp}
\usepackage{cite}
\usepackage{hyperref}
\usepackage{subfigure}
\usepackage{tabularx}
\usepackage{graphicx}
\usepackage{algorithmic}
\usepackage{colortbl}
\usepackage{multirow}
\usepackage{hhline}
\usepackage[table,xcdraw]{xcolor}
\usepackage{url}
\begin{document}
%
\title{On the Self-Repair Role of Astrocytes in STDP Enabled Unsupervised SNNs}

\author{\IEEEauthorblockN{Mehul Rastogi, Sen Lu, Nafiul Islam, Abhronil Sengupta}


%

\thanks{Manuscript dated September, 2020.}
\thanks{The authors are with the School
of Electrical Engineering and Computer Science, The Pennsylvania State University, University Park,
PA 16802, USA. M. Rastogi is also affiliated with Department of of Computer Science and Information Systems, Birla Institute of Technology and Science Pilani, Goa Campus, Goa 403726, India. E-mail: sengupta@psu.edu.
}}


\maketitle

\begin{abstract}
Neuromorphic computing is emerging to be a disruptive computational paradigm that attempts to emulate various facets of the underlying structure and functionalities of the brain in the algorithm and hardware design of next-generation machine learning platforms. This work goes beyond the focus of current neuromorphic computing architectures on computational models for neuron and synapse to examine other computational units of the biological brain that might contribute to cognition and especially self-repair. We draw inspiration and insights from computational neuroscience regarding functionalities of glial cells and explore their role in the fault-tolerant capacity of Spiking Neural Networks (SNNs) trained in an unsupervised fashion using Spike-Timing Dependent Plasticity (STDP). We characterize the degree of self-repair that can be enabled in such networks with varying degree of faults ranging from 50\% - 90\% and evaluate our proposal on the MNIST and Fashion-MNIST datasets.\end{abstract}


%
\IEEEpeerreviewmaketitle

\section{Introduction}
Neuromorphic computing has made significant strides over the past few years - both from hardware \cite{merolla2014million,davies2018loihi,sengupta2017encoding,singh2020nebula} and algorithmic perspective \cite{neftci2019surrogate,diehl2015unsupervised,sengupta2019going,lu2020exploring}. However, the quest to decode the operation of the brain have mainly focused on spike based information processing in the neurons and plasticity in the synapses. Over the past few years, there has been increasing evidence that glial cells, and in particular astrocytes, play a crucial role in a multitude of brain functions \cite{allam2012computational}. As a matter of fact, astrocytes represent a large proportion of the cell population in the human brain \cite{allam2012computational}. There have been also suggestions that complexity of astrocyte functionality can significantly contribute to the computational power of the human brain. Astrocytes are strategically positioned to ensheath tens of thousands of synapses, axons and dendrites among others, thereby having the capability to serve as a communication channel between multiple components and behave as a sensing medium for ongoing brain activity \cite{chung2015astrocytes}. This has led neuroscientists to conclude that astrocytes play a major role in higher order brain functions like learning and memory, in addition to neurons and synapses. Over the past few years, there have been multiple studies to revise the neuron-circuit model for describing higher order brain functions to incorporate astrocytes as part of the neuron-glia network model \cite{allam2012computational,min2012computational}. These investigations clearly indicate and quantify that incorporating astrocyte functionality in network models influence neuron excitability, synaptic strengthening and, in turn, plasticity mechanisms like Short-Term Plasticity and Long-Term Potentiation, which are important learning tools used by neuromorphic engineers. 

The key distinguishing factors of our work against prior efforts can be summarized as follows:

\noindent \textbf{(i)} While recent literature reports astrocyte computational models and their impact on fault-tolerance and synaptic learning \cite{allam2012computational,min2012computational,wade2012self,gordleeva2012bi,de2012computational}, the studies have been mostly confined to small scale networks. This work is a first attempt to explore the self-repair role of astrocytes at scale in unsupervised SNNs in standard visual recognition tasks.

\noindent \textbf{(ii)} In parallel, there is a long history of implementing astrocyte functionality in analog and digital CMOS implementations \cite{ranjbar2015analog,karimi2018neuromorphic,lee2016cmos,amiri2018digital,nazari2015digital,liu2017spanner,irizarry2013cmos}. More recently, emerging physics in post-CMOS technologies like spintronics are also being leveraged to mimic glia functionalities at a one-to-one level \cite{garg2020emulation}. However, the primary focus has been on a brain-emulation perspective, i.e. implementing astrocyte computational models with high degree of detail in the underlying hardware. We explore the aspects of astrocyte functionality that would be relevant to self-repair in the context of SNN based machine learning platforms and evaluate the degree of bio-fidelity required.

\noindent \textbf{(iii)} While Refs. \cite{locally_connected_snn,modified_d&c_archiecture} discusses impact of faults in unsupervised STDP enabled SNNs, self-repair functionality in such networks have not been studied previously. 

While neuromorphic hardware based on emerging post-CMOS technologies \cite{jackson2013nanoscale,kuzum2011nanoelectronic,jo2010nanoscale,ramakrishnan2011floating,sengupta2017encoding} have made significant advancements to reduce the area and power efficiency gap of Artificial Intelligence (AI) systems, such emerging hardware are characterized by a host of non-idealities which has greatly limited its scalability. Our work provides motivation toward autonomous self-repair of such faulty neuromorphic hardware platforms. The efficacy of our proposed astrocyte enabled self-repair process is measured by the following steps: \textbf{(i)} Training SNNs using unsupervised STDP learning rules in networks equipped with lateral inhibition and homeostasis, \textbf{(ii)} Introducing ``faults"\footnote{Note that ``faults" are disjoint from the concept of ``dropout" \cite{srivastava2014dropout} used in neural network training. In dropout, neurons are randomly deleted (along with their connections) only during training to avoid overfitting. In contrast, faults in our work refer to static non-ideal stuck at zero synaptic connections present during both the training and inference stages.} in the trained weight maps by setting a randomly chosen subset of the weights to zero and \textbf{(iii)} Implementing self-repair by re-training the faulty network with astrocyte functionality augmented STDP learning rules. We also compare our proposal with sole STDP based re-training strategy and substantiate our results on the MNIST and F-MNIST datasets.

\section{Materials and Methods}

\subsection{Astrocyte Preliminaries}

\begin{figure}[b]
\centering
\subfigure[]{\includegraphics[width=0.24\textwidth]{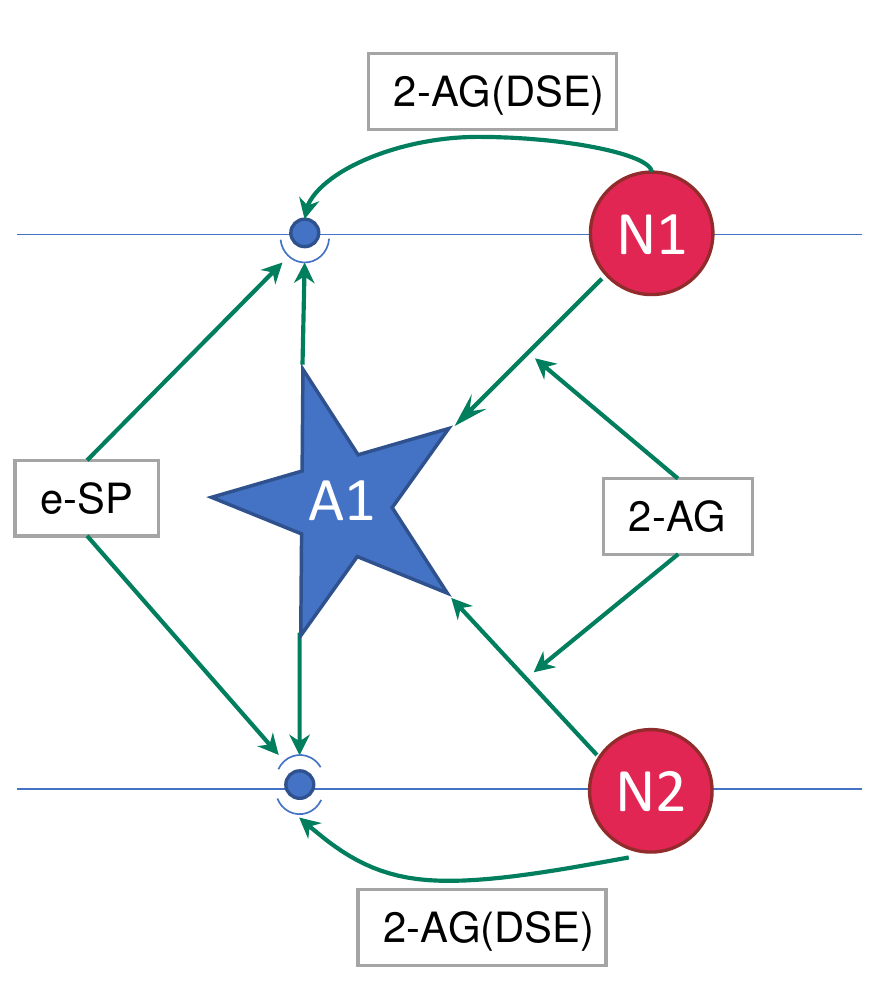}}
\subfigure[]{\includegraphics[width=0.24\textwidth]{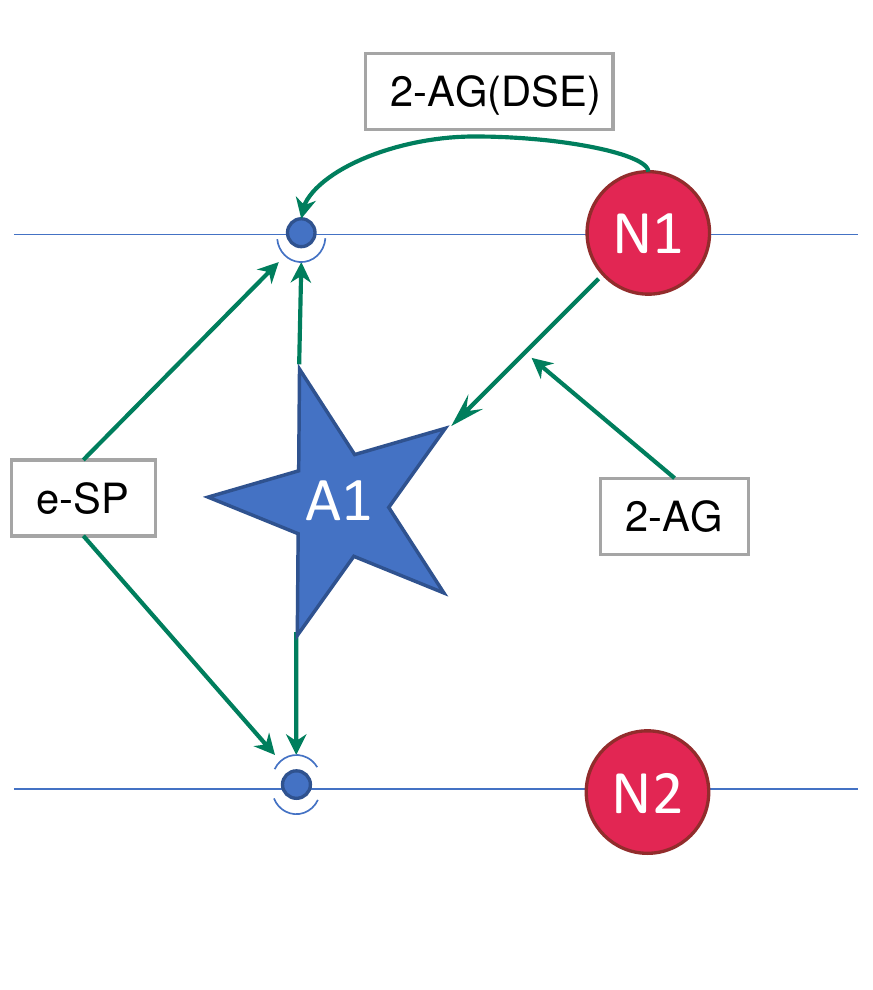}}
\caption{(a) Network with no faults, (b) Network with fault occurring in synapse associated with neuron N2 \cite{wade2012self}. 2-AG is local signal associated with each synapse while e-SP is a global signal. A1 is the astrocyte.}
\label{fig:astrocyte}
\end{figure}

In addition to astrocyte mediated meta-plasticity for learning and memory \cite{wade2012self,volman2007astrocyte,nadkarni2004dressed,nadkarni2007modeling}, there has been indication that retrograde signalling via astrocytes probably underlie self-repair in the brain. Computational models  demonstrate that when faults occur in synapses corresponding to a particular neuron, indirect feedback signal (mediated through retrograde signalling by the astrocyte via endocannabinoids, a type of retrograde messenger) from other neurons in the network implements repair functionality by increasing the transmission probability across all healthy synapses for the affected neuron, thereby restoring the original operation \cite{wade2012self}. Astrocytes modulate this synaptic transmission probability (PR) through two feedback signalling pathways: direct and indirect, responsible for synaptic depression (DSE) and potentiation (e-SP) respectively. Multiple astrocyte computational models \cite{wade2012self,volman2007astrocyte,nadkarni2004dressed,nadkarni2007modeling} describe the interaction of astrocytes and neurons via the tripartite synapse where the astrocyte's sensitivity to 2-arachidonyl glycerol (2-AG), a type of endocannabinoid, is considered. Each time a post synaptic neuron fires, 2-AG is released from the post synaptic dendrite and can be described as:
\begin{equation} \label{eq:1}
    \dfrac{d(\textrm{AG})}{dt}=\dfrac{-\textrm{AG}}{\tau_{\textrm{AG}}}+r_{\textrm{AG}}\delta(t-t_{\textrm{sp}})
\end{equation}
where, AG is the quantity of 2-AG, $\tau_{\textrm{AG}}$ is the decay rate of 2-AG, $r_{\textrm{AG}}$ is the 2-AG production rate and $t_{\textrm{sp}}$ is the time of the post-synaptic spike.

The 2-AG binds to receptors (CB1Rs) on the astrocyte process and instigates the generation of $\textrm{IP}_3$, which subsequently binds to $\textrm{IP}_3$ receptors on the Endoplasmic Reticulum (ER) to open channels that allow the release of $\textrm{Ca}^{2+}$. It is this increase in cystolic $\textrm{Ca}^{2+}$ that causes the release of gliotransmitters into the synaptic cleft that is ultimately responsible for the indirect signaling. The Li-Rinzel model \cite{li1994equations} uses three channels to describe the $\textrm{Ca}^{2+}$ dynamics within the astrocyte: $J_{\textrm{pump}}$ models how $\textrm{Ca}^{2+}$ is pumped into the ER from the cytoplasm via the Sarco-Endoplasmic-Reticulum $\textrm{Ca}^{2+}$-ATPase (SERCA) pumps, $J_{\textrm{leak}}$ describes $\textrm{Ca}^{2+}$ leakage into the cytoplasm and $J_{\textrm{chan}}$ models the opening of $\textrm{Ca}^{2+}$ channels by the mutual gating of $\textrm{Ca}^{2+}$ and $\textrm{IP}_3$ concentrations. The $\textrm{Ca}^{2+}$ dynamics is thus given by:
\begin{equation}
\dfrac{d\textrm{Ca}^{2+}}{dt}=J_{\textrm{chan}}+J_{\textrm{leak}}-J_{\textrm{pump}}
\end{equation}
The details of the equations and their derivations can be obtained from Refs. \cite{wade2012self} and \cite{de2009glutamate}.

The intracellular astrocytic calcium dynamics control the glutamate release from the astrocyte which drives e-SP. This release can be modelled by:
\begin{equation}
    \dfrac{d(\textrm{Glu})}{dt}=\dfrac{-\textrm{Glu}}{\tau_{\textrm{Glu}}}+r_{\textrm{Glu}}\delta(t-t_{\textrm{Ca}})
\end{equation}
where, Glu is the quantity of glutamate, $\tau_{\textrm{Glu}}$ is the glutamate decay rate, $r_{\textrm{Glu}}$ is the glutamate production rate and $t_{\textrm{Ca}}$ is the time of the $\textrm{Ca}^{2+}$ threshold crossing. To model e-SP:
\begin{equation} \label{eq:15}
    \tau_{\textrm{eSP}}\dfrac{d(\textrm{eSP})}{dt}=-\textrm{eSP} + m_\textrm{eSP}\textrm{Glu}(t)
\end{equation}
where, $\tau_{\textrm{eSP}}$ is the decay rate of e-SP and $m_{\textrm{eSP}}$ is a scaling factor. Eq. (4) substantiates that the level of e-SP is dependent on the quantity of glutamate released by the astrocyte.

\begin{figure}[!t]
\centering
\subfigure[]{\includegraphics[trim = 0cm 19cm 0cm 0cm, clip, width=0.48\textwidth]{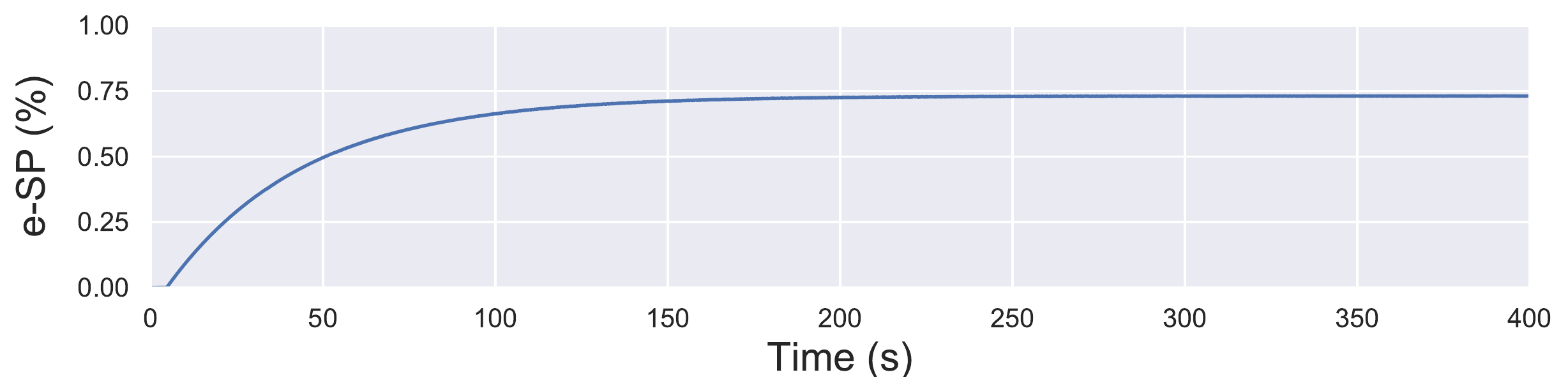}}\linebreak
\subfigure[]{\includegraphics[trim = 0cm 12.7cm 0cm 6.5cm, clip, width=0.48\textwidth]{figures/PR_fault70.pdf}}\linebreak
\subfigure[]{\includegraphics[trim = 0cm 6.5cm 0cm 12.6cm, clip, width=0.48\textwidth]{figures/PR_fault70.pdf}}\linebreak
\subfigure[]{\includegraphics[trim = 0cm 0.3cm 0cm 18.8cm, clip, width=0.48\textwidth]{figures/PR_fault70.pdf}}
\caption{Simulation results of the network in Fig. 1 using the computational model of astrocyte mediated self-repair from \cite{wade2012self}. Total simulation time is 400s. At 200s, faults are introduced in 70\% of the synapses connected to N2. All the synapses have PR($t_0$)=0.5. (a) e-SP of N1 and N2. It is the same for both N1 and N2 since e-SP is a global function, (b) DSE of N1 and N2. It is different for each neuron as it is dependent upon the neuron output. At 200s, after the introduction of the faults in N2, only DSE of N2 changes, (c) PR of different types of synapses connected to N1 and N2, and (d) Firing rate of neurons N1 and N2.}
\label{fig:network_simulation}
\end{figure}

The released 2-AG also binds directly to pre-synpatic CB1Rs (direct signaling). A linear relationship is assumed between DSE and the level of 2-AG released by the post-synaptic neuron as:
\begin{equation}
\textrm{DSE} = -\textrm{AG} \times K_\textrm{AG}
\end{equation}
where, AG is the amount of 2-AG released by the post-synaptic neuron and is found from Eq. (1) and $K_{\textrm{AG}}$ is a scaling factor.
The PR associated with each synapse is given by the following equation:  
\begin{equation}
\textrm{PR}(t)= \textrm{PR}(t_{0}) + \textrm{PR}(t_{0})\times \left(\frac{\textrm{DSE}(t)+\textrm{eSP}(t)}{100}\right)
\label{eqn:adaptive-PR}
\end{equation}
where, PR($t_0$) is the initial PR of the synapses, e-SP and DSE are given by Eq. (4) and (5) respectively. In the computational models, the effect of DSE is local to the synapses connected to a particular neuron whereas all the tripartite synapses connected to the same astrocyte receive the same e-SP. Under no-fault condition, the DSE and e-SP reach a dynamic equilibrium where the PR is unchanged over time, resulting in a fixed firing rate for the neurons. When a fault occurs, this balance subsides and the PR changes according to Eq. (6) to restore the firing rate to its previous value. To showcase this effect consider for instance, Fig. \ref{fig:astrocyte} where a simple SNN with two post-synaptic neurons is depicted. Let us assume that each post-neuron receives input spikes from 10 pre-neurons. The initial PR of the synapses were set to 0.5. Fig. \ref{fig:astrocyte}(a) is the case with no faults, while in Fig. \ref{fig:astrocyte}(b), faults have occurred after some time in $70\%$ of the synapses associated with post-neuron N2 (Fig. \ref{fig:network_simulation}). Note, here ``faults" imply that the synapses do not take part in transmission of the input spikes i.e. have a PR of 0. This results in a drop of the firing frequency associated with N2 while operation of N1 is not impacted. Thus, the amount of 2-AG released by N2 decreases, which increases DSE and in turn increases the PR of the associated synapses of N2 where no faults have occurred. Hence, we observe in Fig. \ref{fig:network_simulation}(d) that the increased PR recovers the firing rate and approaches the ideal firing frequency. Note that the degree of self-recovery, i.e. the difference between the recovered and ideal frequency is a function of the fault probability. The simulation conditions and parameters for the modelling are based on Ref. \cite{wade2012self}. Interested readers are directed to Ref. \cite{wade2012self} for an extensive discussion on the astrocyte computational model and the underlying processes governing the retrograde signalling.

\begin{figure}[b]
\centering
    \includegraphics[width=0.51\textwidth]{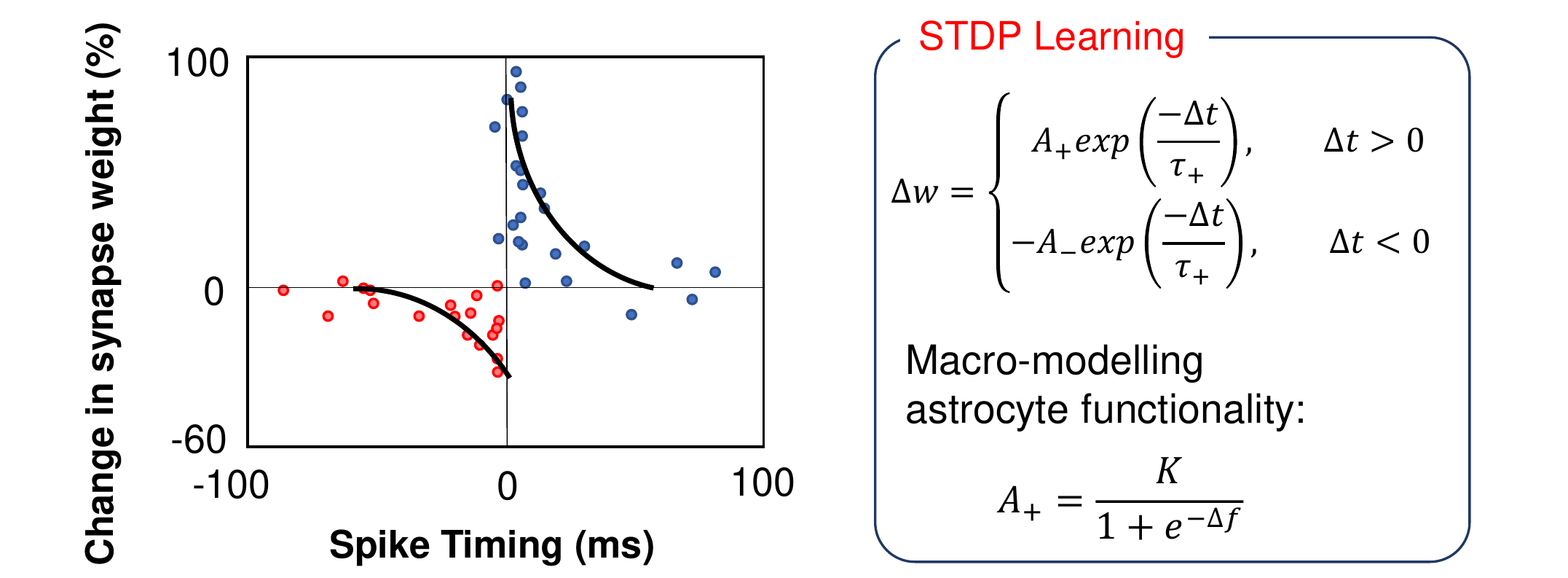}
\caption{In the above equations, the STDP learning window height is a non-linear increasing function of the deviation $\Delta f$ from the ideal firing frequency of the post-neuron.}
\label{fig:bcm}
\end{figure}

A key question that we have attempted to address in this work is the computational complexity at which we require to model the feedback mechanism to implement autonomous repair in such self-learning networks. Simplifying the feedback modelling would enable us to implement such functionalities by efficient hardware primitives. For instance, the core functionality of astrocyte self-repair occurs in conjunction with STDP based learning in synapses. Fig. \ref{fig:bcm} shows a typical STDP learning rule where the change in synaptic weight varies exponentially with the spike time difference between the pre- and post-neuron \cite{liu2018exploring}, according to measurements performed in rat glutamatergic synapses \cite{bi2001synaptic}. Typically, the height of the STDP weight update for potentiation/depression is constant ($A_+$/$A_-$). However, astrocyte mediated self-repair suggests that the weight update should be a function of the firing rate of the post-neuron \cite{liu2018exploring}. Assuming the fault-less expected firing rate of the post-neuron to be $f_{ideal}$ and the non-ideal firing rate to be $f$, the synaptic weight update window height should be a function of $\Delta f=f_{ideal}-f$. The concept has been explained further in Fig. \ref{fig:bcm}  and is also in accordance with Fig. \ref{fig:network_simulation} where the PR increase after fault introduction varies in a non-linear fashion over time and eventually stabilizes once the self-repaired firing frequency approaches the ideal value. The functional dependence is assumed to be that of a sigmoid function – indicating that as the magnitude of the fault, i.e. deviation in the ideal firing frequency of the neuron increases, the height of the learning window increases in proportion to compensate for the fault \cite{liu2018exploring}. Note that the term ``fault" for the machine learning workloads, described herein, refers to synaptic weights (symbolizing PR) stuck at zero. Therefore, with increasing amount of synaptic faults, $f << f_{ideal}$, thereby increasing the STDP learning window height significantly. During the self-healing process, the frequency deviation gradually reduces and thereby the re-learning rate also becomes less pronounced and finally saturates once the ideal frequency is reached. While our proposal is based on Ref. \cite{liu2018exploring}, prior work has been explored in the context of a prototype artificial neural network with only 4 input neurons and 4 output neurons. Extending the framework to an unsupervised SNN based machine learning framework therefore requires significant explorations, highlighted next. 

\vspace{-2mm}

\subsection{Neuron Model and Synaptic Plasticity}
We utilized the Leaky Integrate and Fire (LIF) spiking neuron model in our work. The temporal LIF neuron dynamics are described as,
\begin{equation}
\begin{split}
\tau_{mem} \frac{\partial v(t)}{\partial t} =  -v(t) + v_{rest} + I(t)
\end{split}
\label{eqn:lif}
\end{equation}
where, $v(t)$ is the membrane potential, $\tau_{mem}$ is the membrane time constant, $v_{rest}$ is the resting potential and $I(t)$ denotes the total input to the neuron at time $t$. The weighted summation of synaptic inputs is represented by $I(t)$. When the neuron's membrane potential crosses a threshold value, $v_{th}(t)$, it fires an output spike and the membrane potential is reset to $v_{reset}$. The neuron's membrane voltage is fixed at the reset potential for a refractory period, $\delta_{ref}$, after it spikes during which it does not receive any inputs.

In order to ensure that single neurons do not dominate the firing pattern, homeostasis \cite{diehl2015unsupervised} is also implemented through an adaptive thresholding scheme. The membrane threshold of each neuron is given by the following temporal dynamics,
\begin{equation}
\begin{split}
& v_{th} (t) = \theta_0 + \theta(t)\\
& \tau_{theta} \frac{\partial \theta(t)}{\partial t} = - \theta(t)
\end{split}
\label{eqn:adaptive-threshold}
\end{equation}
where, $\theta_0 > v_{rest},v_{reset}$ and is a constant. $\tau_{theta}$ is the adaptive threshold time constant. The adaptive threshold, $\theta(t)$ is increased by a constant quantity $\theta_{+}$, each time the neuron fires, and decays exponentially according to the dynamics in Equation \ref{eqn:adaptive-threshold}.

A trace \cite{trace_stdp} based synaptic weight update rule was used for the online learning process \cite{diehl2015unsupervised,locally_connected_snn}. The pre and post-synaptic traces are given by $x_{pre}$ and $x_{post}$ respectively. Whenever the pre (post) - synaptic neuron fires, the variable $x_{pre}$ ($x_{post}$) is set to 1, otherwise it decays exponentially to 0 with spike trace decay time constant, $\tau_{trace}$. The STDP weight update rule is characterized by the following dynamics, 
\begin{equation}
\Delta w = \begin{cases} 
\eta_{post} * x_{pre}  & \mbox{on post-synaptic spike}\\
-\eta_{pre} * x_{post} & \mbox{on pre-synaptic spike}
\end{cases}
\label{eqn:trace-stdp-model}
\end{equation}
where, $ \eta_{pre} / \eta_{post} $ denote the learning rates for pre-synaptic / post-synaptic updates respectively. The weights of the neurons are bounded in the range of $[0,w_{max}]$. It is worth mentioning here that the sum of the weights associated with all post-synaptic neurons is normalized to a constant factor, $w_{norm}$ \cite{locally_connected_snn}.

\subsection{Network Architecture}
\begin{figure}[h]
\centering
    \includegraphics[width=0.4\textwidth]{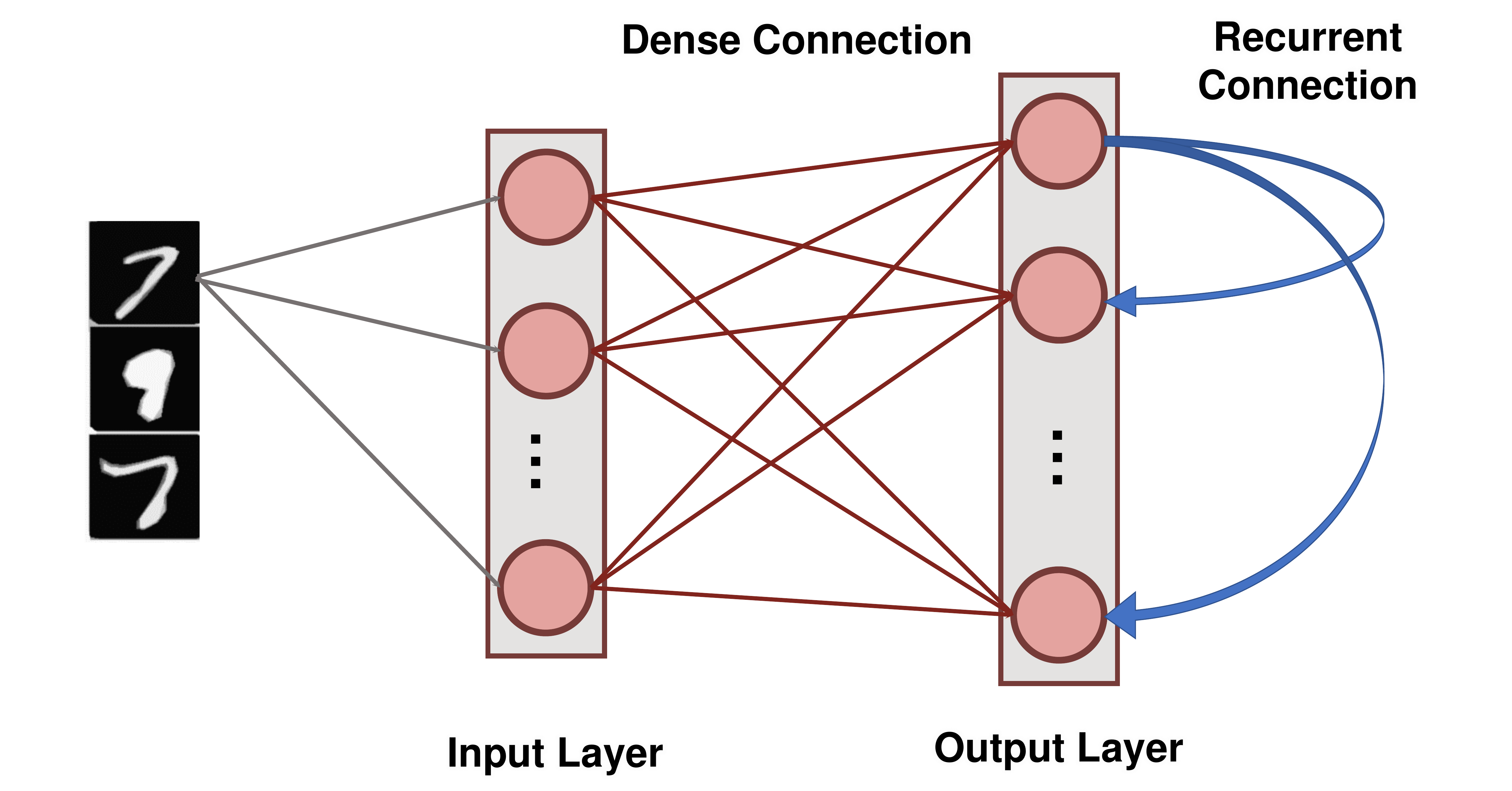}
\caption{The single layer SNN architecture with lateral inhibition and homeostasis used for unsupervised learning.}
\label{fig:network}
\end{figure}
Our SNN based unsupervised machine learning framework is based on single layer architectures inspired from cortical microcircuits \cite{diehl2015unsupervised}. Fig. \ref{fig:network} shows the network connectivity of spiking neurons utilized for pattern-recognition problems. Such a network topology has been shown to be efficient in several pattern-recognition problems, such as digit recognition \cite{diehl2015unsupervised} and sparse encoding \cite{knag2015sparse}. The SNN, under consideration, has an Input Layer with the number of neurons equivalent to the dimensionality of the input data. Input neurons generate spikes by converting each pixel in the input image to a Poisson spike train whose average firing frequency is proportional to the pixel intensity. This layer connects in an all-to-all fashion to the Output Layer through excitatory synapses. The Output layer has $n_{neurons}$ LIF neurons characterized by homeostasis functionality. It also has static (constant weights) recurrent inhibitory synapses with weight values, $w_{recurrent}$, for lateral inhibition to achieve soft Winner-Take-All (WTA) condition. Each neuron in the Output Layer has an inhibitory connection to all the neurons in that layer except itself. Trace-based STDP mechanism is used to learn the weights of all synapses between the Input and Output Layers. The neurons in the Output Layer are assigned classes based on their highest response (spike frequency) to input training patterns \cite{diehl2015unsupervised}.

\subsection{Challenges and Astrocyte Augmented STDP (A-STDP) Learning Rule Formulation}
\begin{figure}[h]
\centering
    \includegraphics[width=0.45\textwidth]{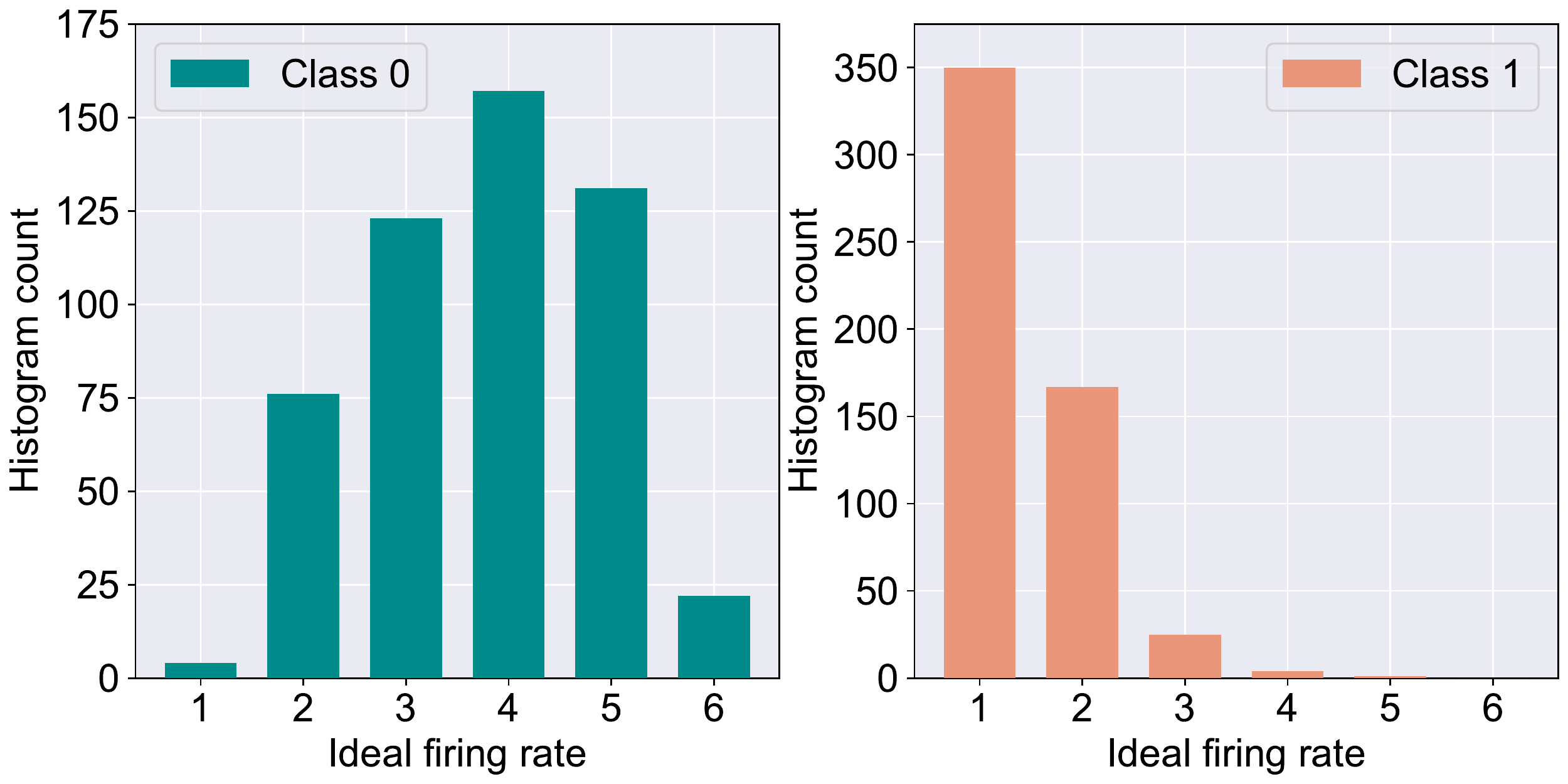}
\caption{Histogram count of the ideal firing rate of neurons responding to digit `0' versus digit `1' (measured from 5000 test examples of the MNIST dataset).}
\label{fig:histogram}
\end{figure}

One of the major challenges in extending the astrocyte based macro-modelling in such self-learning networks lies in the fact that the ideal neuron firing frequency is a function of the specific input class the neuron responds to. This is substantiated by Fig. \ref{fig:histogram} which depicts the histogram distribution of the ideal firing rate of the wining neuron in the fault-less network. Further, due to sparse neural firing, the total number of output spikes of the winning neurons over the inference window is also small, thereby limiting the amount of information (number of discrete levels) that can be encoded in the frequency deviation, $\Delta f$. This leads to the question: \textit{Can we utilize another surrogate signal that gives us information about the degree of self-repair occurring in the network over time while being independent of the class of the input data?}

While the above challenge is related to the process of reducing the STDP learning window over time, we observed that using sole STDP learning or with a constant enhanced learning rate consistently reduced the network accuracy over time (Fig. \ref{fig:comparison}).  Fig. \ref{fig:weight_map_stdp} also depicts that normal STDP retraining with faulty synapses slowly loses their learnt representations over time. Re-learning all the healthy synaptic weights uniformly using STDP with an enhanced learning rate should at least result in some accuracy improvement for the initial epochs of re-training, even if the modulation of learning window height over time is not incorporated in the self-repair framework. The degradation of network accuracy starting from the commencement of the retraining process signified that some additional factors may have been absent in the astrocyte functionality macro-modelling process, which is independent from the above challenge of modulating the temporal behavior of the STDP learning window. 

\begin{figure}[h]
\centering
\subfigure[]{\includegraphics[trim=0cm 19cm 0cm 0cm,clip,width=0.48\textwidth]{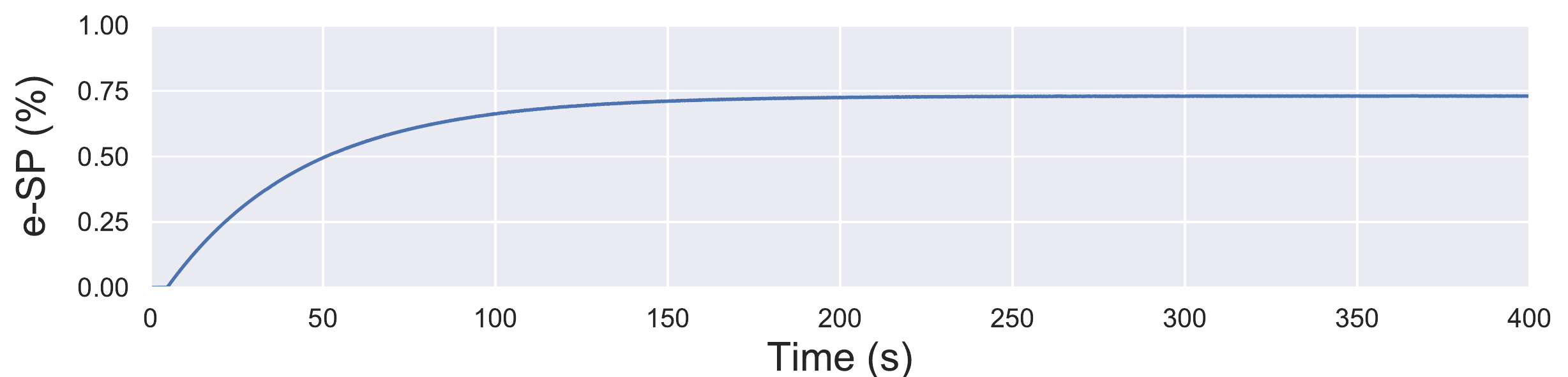}}\linebreak
\subfigure[]{\includegraphics[trim=0cm 12.7cm 0cm 6.5cm,clip,width=0.48\textwidth]{figures/self_rep.pdf}}\linebreak
\subfigure[]{\includegraphics[trim=0cm 6.5cm 0cm 12.6cm,clip,width=0.48\textwidth]{figures/self_rep.pdf}}\linebreak
\subfigure[]{\includegraphics[trim=0cm 0.3cm 0cm 18.8cm,clip,width=0.48\textwidth]{figures/self_rep.pdf}}
\caption{Simulation results of the network in Fig. 1 using the computational model of \cite{wade2012self} with synapses having different initial PR values. Total simulation time is 400s. At 200s, faults are introduced in 8 synapses with high initial PR connected to N2. (a) e-SP of N1 and N2, (b) DSE of N2, (c) PR of the 3 types of synapses connected to N2 (orange: healthy synapse with PR($t_0$)=0.5, green: healthy synapse with PR($t_0$)=0.1 and blue: faulty synapse with PR($t_0$)=0.5 till 200s and PR($t_0$)=0 afterwards) and (d) Firing rate of neuron N2}.
\label{fig:wade_PR_simulation}
\end{figure}

\begin{figure}[h]
\centering
    \includegraphics[width=0.4\textwidth]{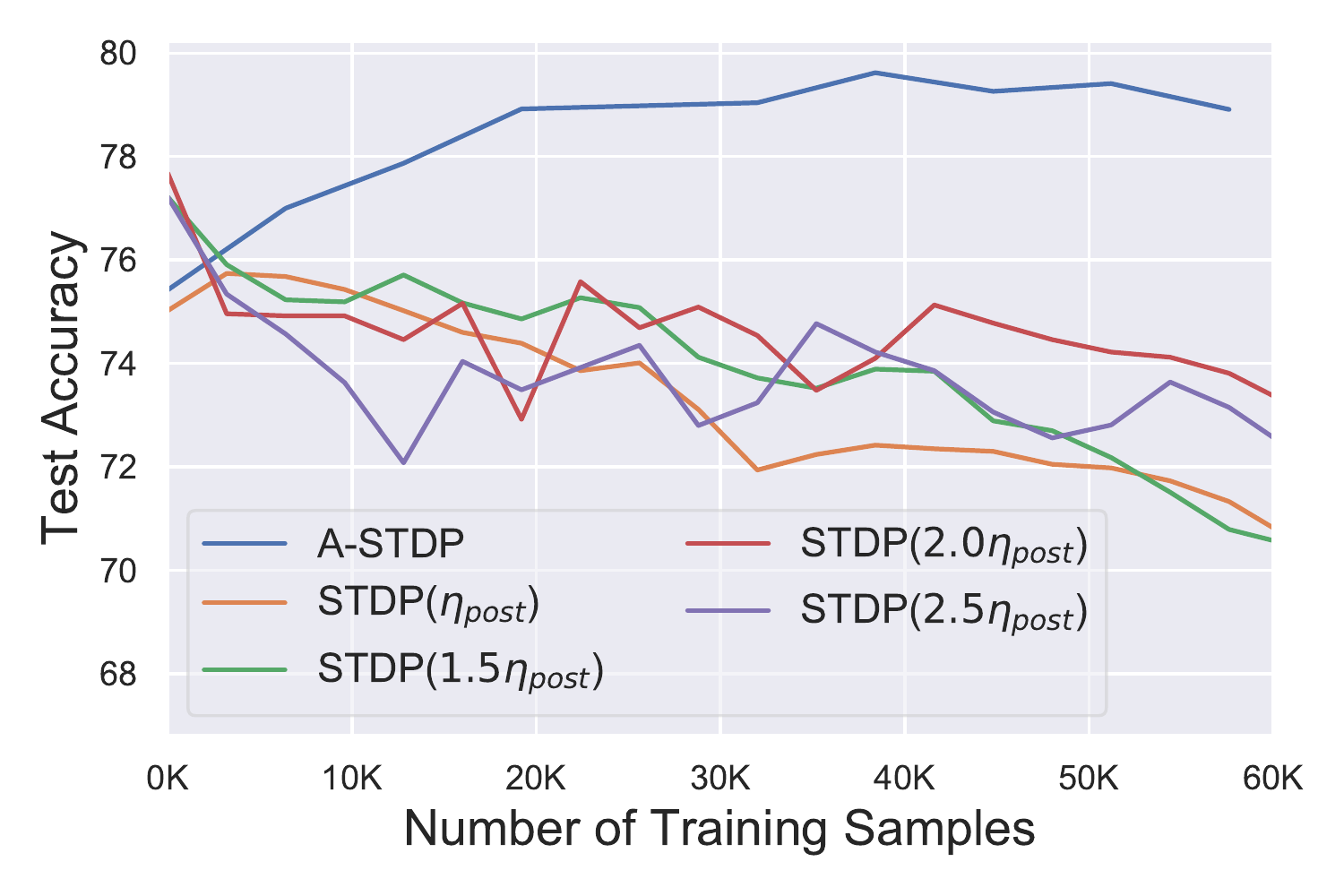}
\caption{Test accuracy of a 225 neuron network on the MNIST dataset with 70\% faulty connections with normal and enhanced learning rates during STDP re-training process. Re-training with A-STDP rule is also depicted.}
\label{fig:comparison}
\end{figure}

\begin{figure}[!t]
\centering
50\% Fault Probability \linebreak
\medskip
\subfigure[Baseline Network]{\includegraphics[trim={2.5cm 0 0 0}, clip, width=0.24\textwidth]{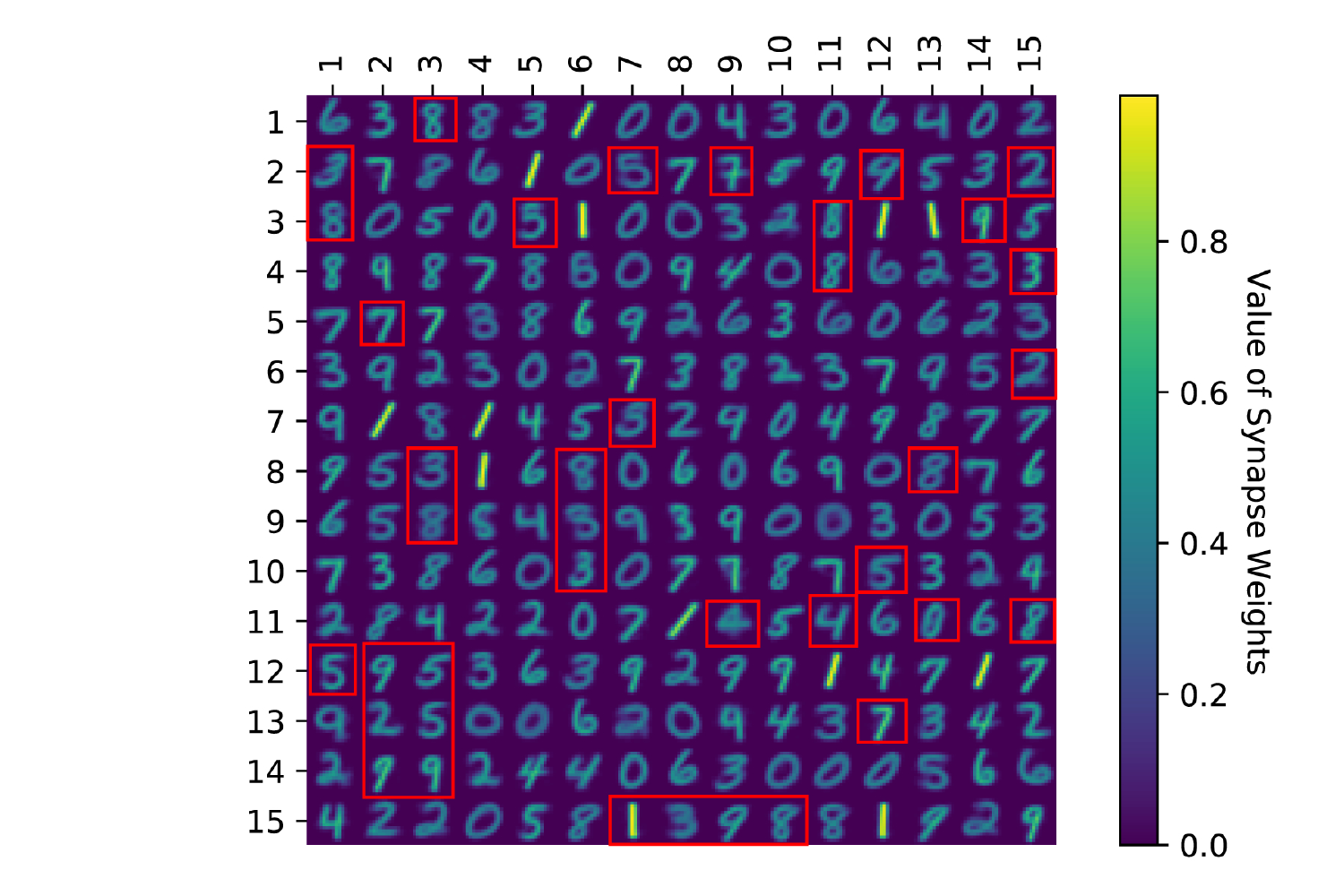}}
\subfigure[After STDP Re-learning]{\includegraphics[trim={2.5cm 0 0 0}, clip, width=0.24\textwidth]{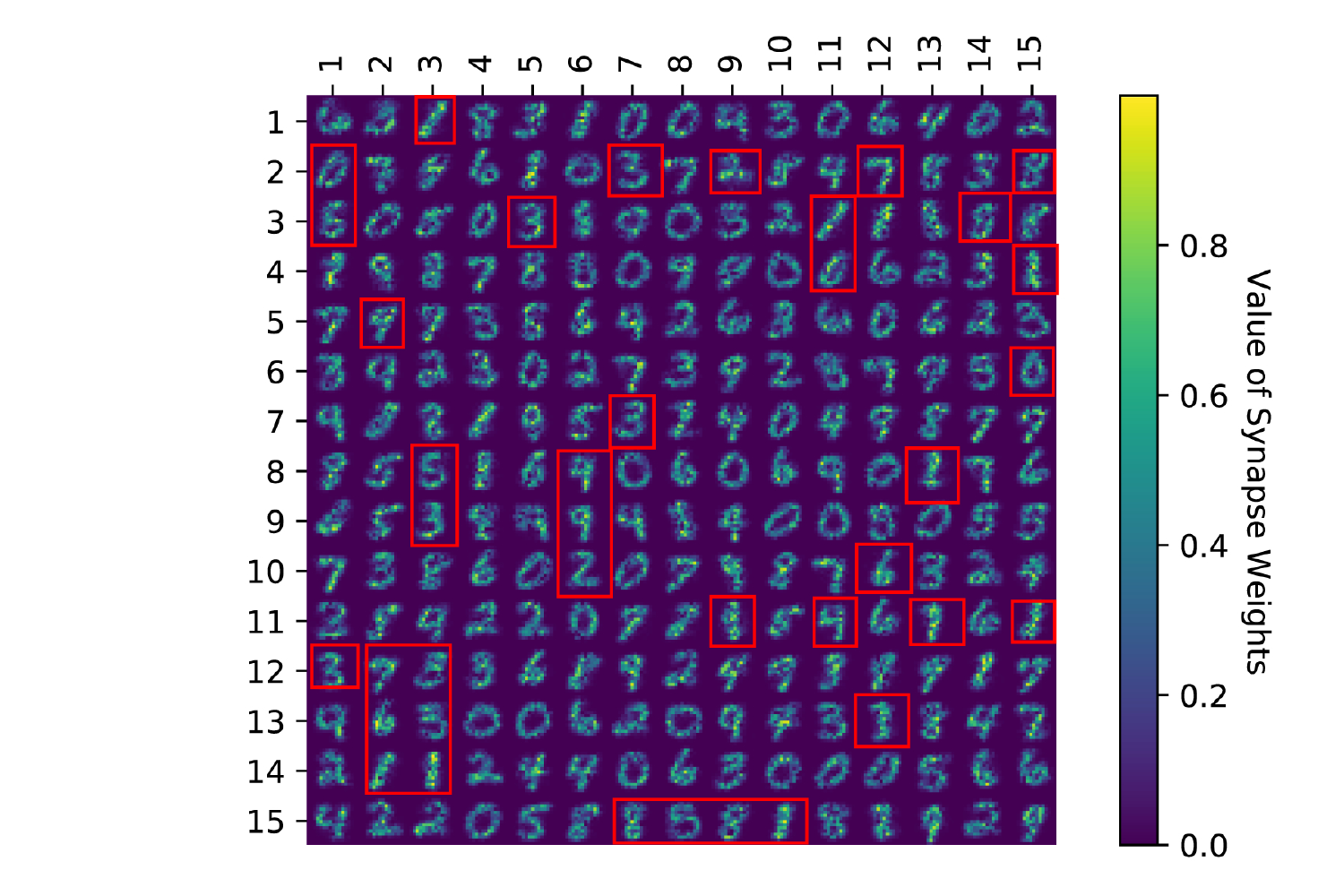}}
80\% Fault Probability  \linebreak
\subfigure[Baseline Network]{\includegraphics[trim={2.5cm 0 0 0}, clip, width=0.24\textwidth]{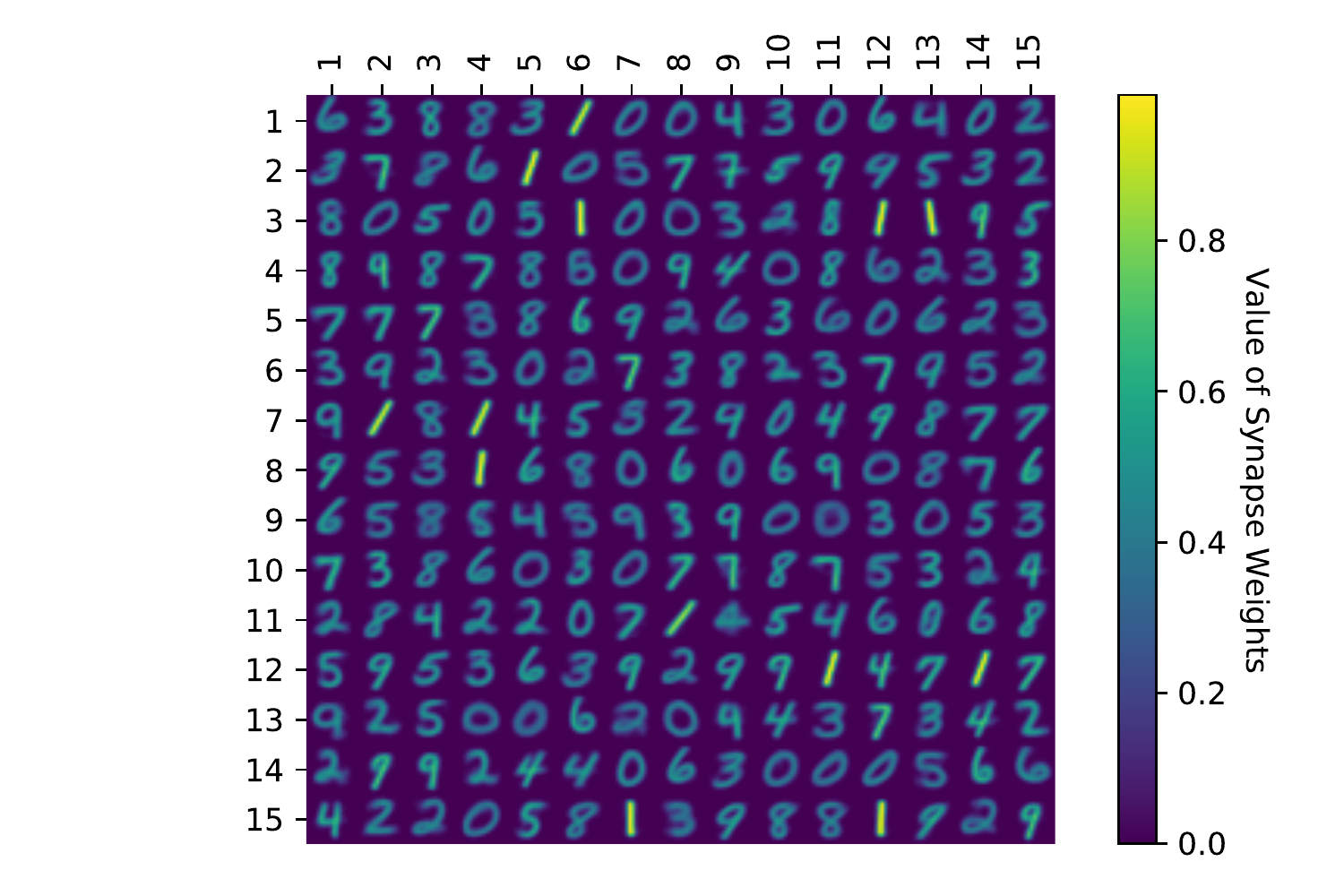}}
\subfigure[After STDP Re-learning]{\includegraphics[trim={2.5cm 0 0 0}, clip, width=0.24\textwidth]{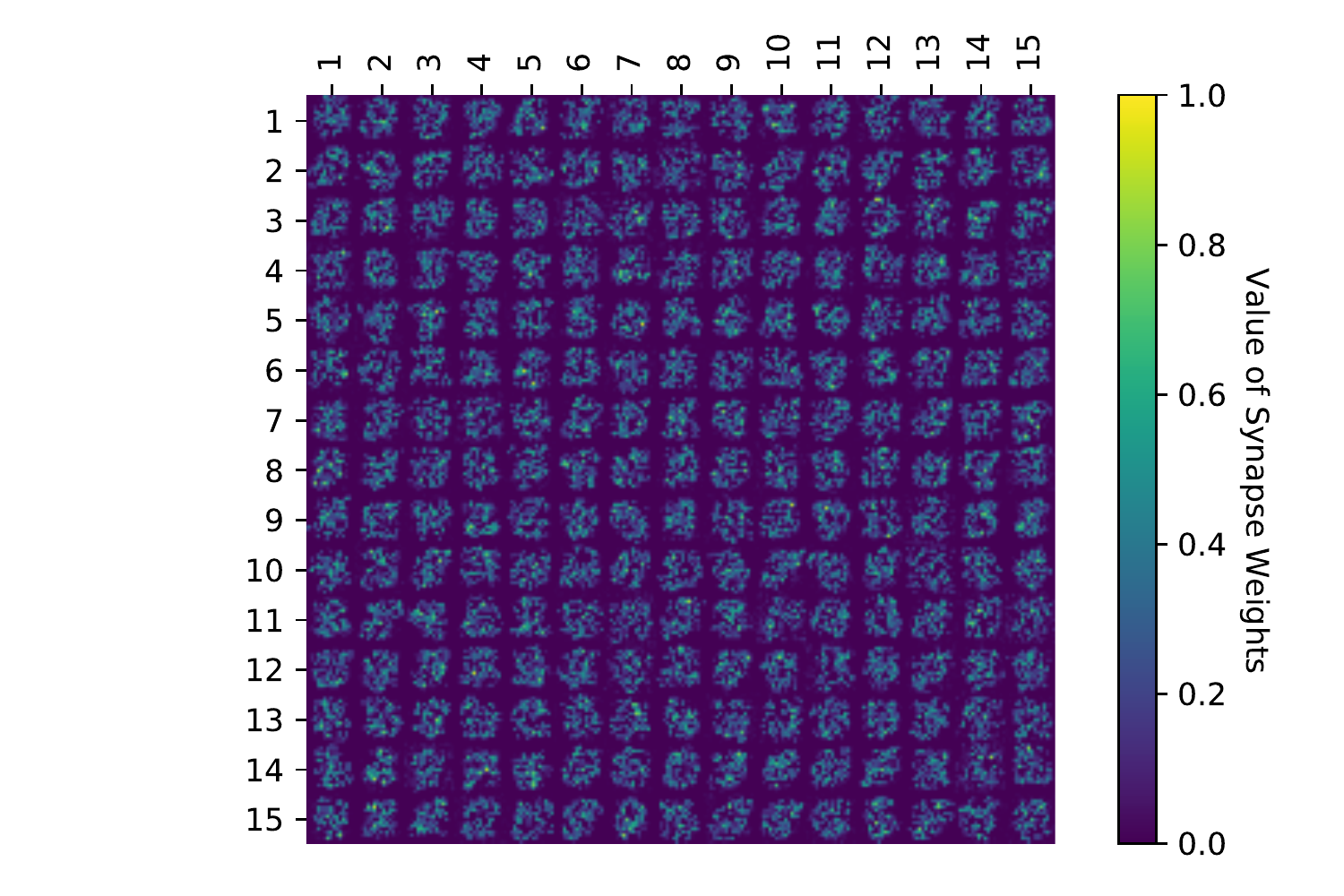}}
\caption{(a)-(d) Learnt weight patterns for 225 neuron network on the MNIST dataset are shown. Re-training the network with sole STDP learning causes distortion of the weight maps (50\% and 80\% fault cases are plotted). The red boxes in (a) and (b) highlight how the the neurons can change association toward a particular class during re-learning thereby forgetting their original learnt representations. Receptive fields of all neurons undergo distortion for the 80\% fault case.}
\label{fig:weight_map_stdp}
\end{figure}

In that regard, we draw inspiration from Eq. 6, where we observe that the initial fault-free value of the PR acts as a scaling factor for the self-repair feedback terms DSE and e-SP. We perform a similar simulation for the network shown in Fig. 1, with each neuron receiving input from 10 synapses. However in this case, we set the initial PR of all of the synapses to 0.5, except one connected to N2; for which the initial PR was set to 0.1. In other words, 9 of the synapses connected to N2 have a PR($t_0$)=0.5, while for one PR($t_0$)=0.1. The lower initial PR value symbolizes a weaker connection. The network is simulated for 400s and at 200s, the associated PR of 8 of the synapses with higher initial PR are reduced to 0 to signify faulty condition (Fig. \ref{fig:wade_PR_simulation}). We observe that after the introduction of the faults, the PR of the synapses with the higher initial PR value is enhanced greatly compared to the one with the lower initial PR.  This leads us to the conclusion that synapses that play a greater role in postsynaptic firing also play a greater role in the self-repair process compared to other synapses.

Since our unsupervised SNN is characterized by analog synaptic weights in the range of $[0,w_{max}]$, we hypothesized that this characteristic might underlie the reason for the accuracy degradation and designed a preferential self-repair learning rule for healthier synapses. This was found to result in significant accuracy improvement during the retraining process (discussed in next section). Our formulated A-STDP learning rule formulation is therefore also guided by the following question: \textit{Can we aggressively increase the healthy synaptic weights during the initial learning epochs which preserves the original representations learnt by the network?} 

\begin{figure}[h]
\centering
   \includegraphics[width=0.4\textwidth]{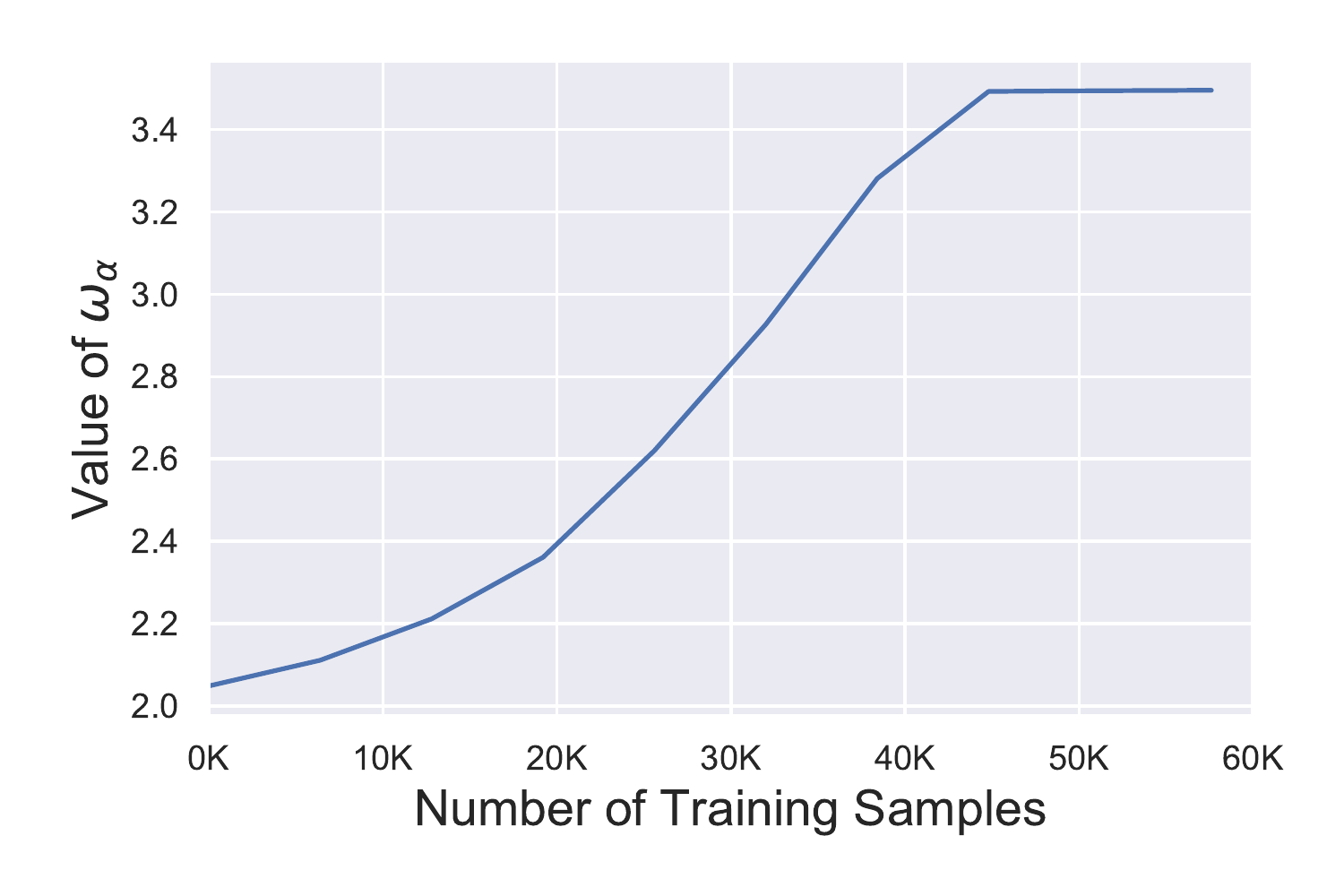}
\caption{Value of $w_\alpha$ (98-th percentile from weight distribution of the entire network) during the self-repair process using A-STDP learning rule for a 225 neuron network on the MNIST Dataset with 80\% faulty connections.}
\label{fig:alpha}
\end{figure}

Driven by the above observations, we formulated our Astrocyte Augmented STDP (A-STDP) learning rule during the self-repair process as,
\begin{equation}
\begin{split}
\Delta w = \begin{cases} 
\eta_{post} * x_{pre}  * \boldsymbol{(w / w_\alpha )^{\sigma}} & \mbox{on post-synaptic spike}\\
-\eta_{pre} * x_{post} & \mbox{on pre-synaptic spike}
\end{cases}
\end{split}
\label{eqn:trace-weighted-stdp-model}
\end{equation}
where, $w_\alpha$ represents the the weight value at the $\alpha$-th percentile of the network and serves as the surrogate signal to guide the retraining process. Fig. \ref{fig:alpha} depicts the temporal behavior of $w_\alpha$ for the 98-th percentile of the weight distribution. After faults are introduced, $w_\alpha$ is significantly reduced and slowly increases over time during the re-learning process. It finally saturates off at the bounded value $w_{max}$. The term $w / w_\alpha$ ensures that the effective learning rate for healthier synapses ($w > w_\alpha$) is much higher than the learning rate for weaker connections ($w < w_\alpha$) while $\sigma$ dictates the degree of non-linearity. Since $w_\alpha$ increases over time, the enhanced learning process also reduces and finally stops once  $w_\alpha$ saturates. It is worth mentioning here that $w_\alpha$, $\sigma$ and $w_{max}$ are hyperparameters for the A-STDP learning rule. All hyperparameter settings and simulation details are presented in the next section. 

\section{Results}

\begin{figure}[htp]
\centering
\subfigure[]{\includegraphics[width=0.24\textwidth]{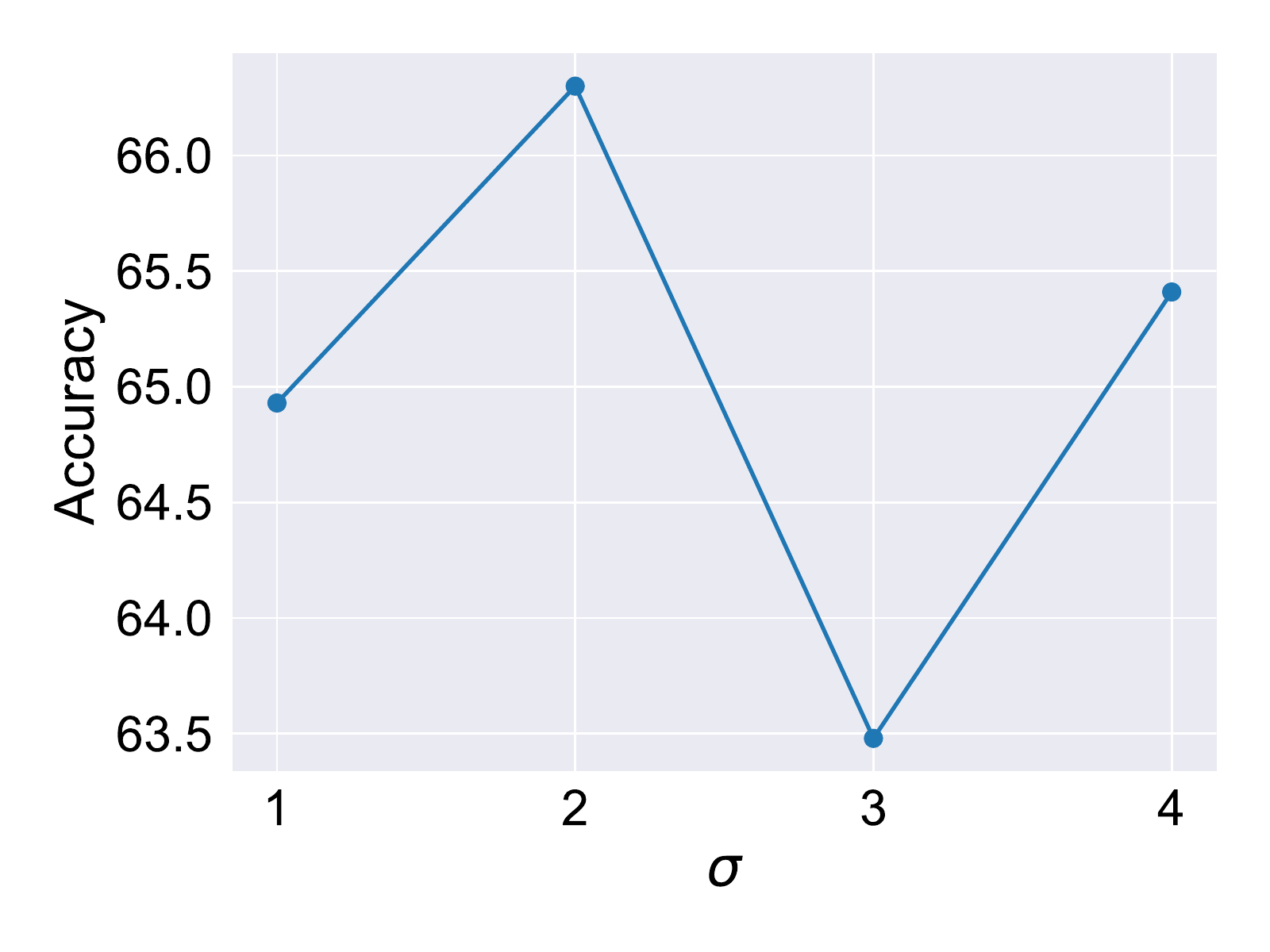}}
\subfigure[]{\includegraphics[width=0.24\textwidth]{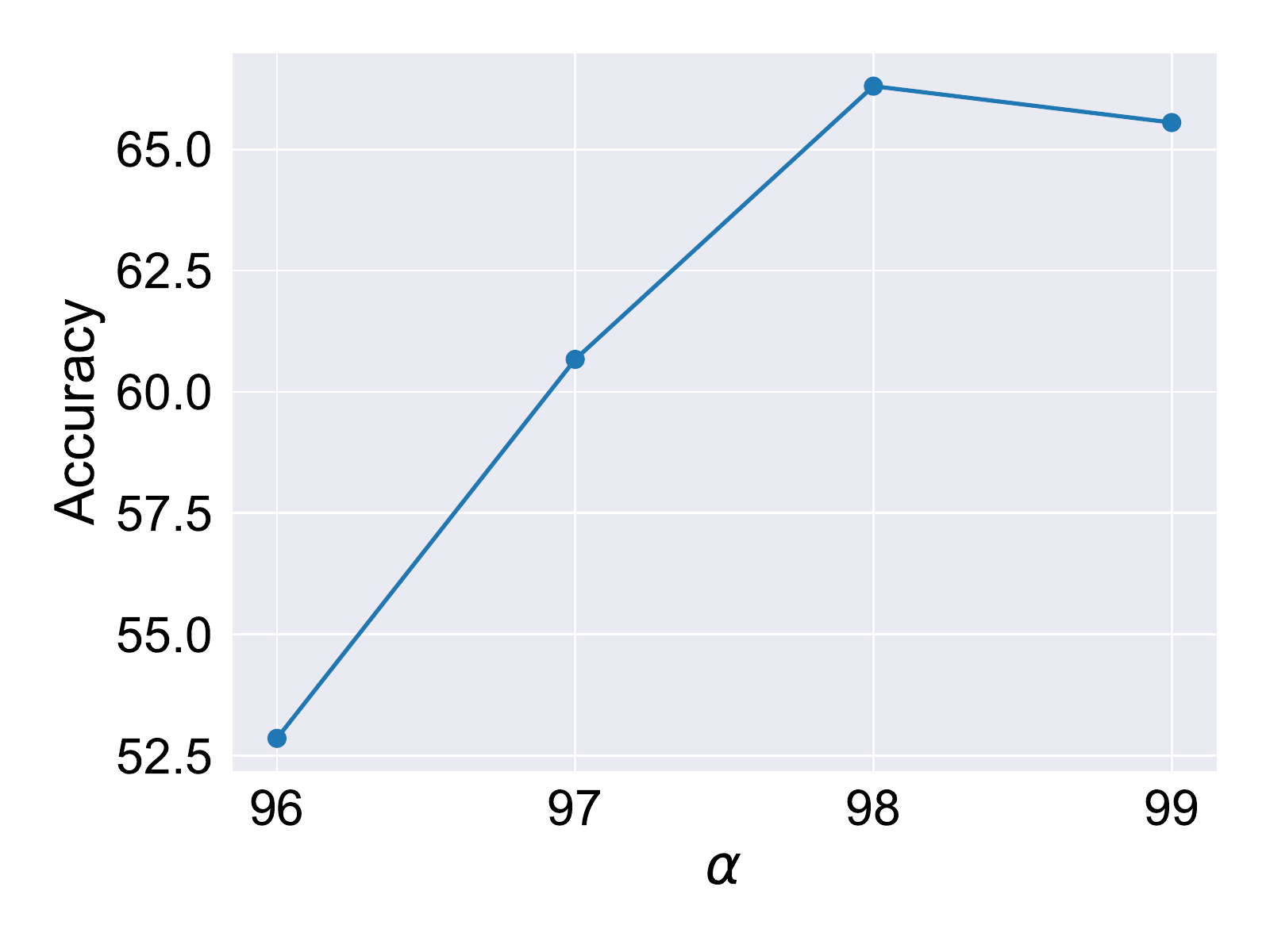}}

\caption{Ablation studies for the hyperparameters (a) $\sigma$ (with fixed $\alpha=98$) and (b) $\alpha$ (with fixed $\sigma=2$) in A-STDP learning rule. }
\label{fig:ablation}
\end{figure}

We evaluated our proposal in the context of unsupervised SNN training on standard image recognition benchmarks under two settings: scaling in network size and scaling in network complexity. We used MNIST \cite{lecun-mnisthandwrittendigit-2010} and Fashion-MNIST \cite{fmnist} datasets for our analysis. Both datasets  contain 28 $\times$ 28 grayscale images of handwritten digits / fashion products (belonging to one of 10 categories) with 60,000 training examples and 10,000 testing examples. All experiments are run in PyTorch framework using a single GPU with a batchsize of 16 images. In addition to standard input pre-processing for generating the Poisson spike train, the images in F-MNIST dataset also undergo Sobel filtering for edge detection before being converted to spike trains. The SNN implementation is done using a modified version of the mini-batch processing enabled SNN simulation framework \cite{saunders2019minibatch} in BindsNET \cite{Hazan_2018}, a PyTorch based package \href{https://github.com/BindsNET/bindsnet}{(Link)}. In addition to dataset complexity scaling, we also evaluated two networks with increasing size (225 and 400 neurons) on the MNIST dataset. For the MNIST dataset, the baseline test accuracy of the ideal network was 89.53\% and 92.02\% respectively. A 400-neuron network was used for the F-MNIST dataset with 77.35\% accuracy. The baseline test accuracies are superior/comparable to prior reported accuracies for unsupervised learning on both datasets. For instance, Ref. \cite{diehl2015unsupervised} reports 87\% accuracy for an STDP trained network with 400 neurons while Ref. \cite{fmnist_acc_unsupervised} reports the best accuracy of 73.1\% for state-of-the-art clustering methods on the F-MNIST dataset. Table \ref{table1} lists the network simulation parameters used in this work. It is worth mentioning here that all hyperparameters were kept unchanged (from their initial values during training) in the self-repair process. We also kept the hyperparameters, $w_\alpha$ and $\sigma$ for the A-STDP rule unchanged for all fault simulations. Fig. \ref{fig:ablation} shows a typical ablation study of the hyperparameters $\alpha$ and $\sigma$. For this study, we trained a 225-neuron network with 90\% faults. We divided the training set into training and validation subsets in the ratio of 5:1 respectively through random sampling. The two accuracy plots shown in Fig. \ref{fig:ablation} are models retrained on the training subset and then evaluated on the new validation set. Further hyperparameter optimizations for different fault conditions can potentially improve the accuracy improvement even further. 

\begin{table}[t]
\caption{Simulation Parameters}
\label{table1}
\center
\begin{tabular}{c c}
\hline \hline
\bfseries Parameters & \bfseries Value\\
\hline
Membrane Time Constant, $\tau_{mem}$ & 100ms  \\
Spike Trace Decay Time Constant, $\tau_{trace}$ & 20ms \\
Resting Potential, $v_{rest}$ & -65mV \\
Threshold Voltage Constant, $\theta_0$ & -52mV \\
Membrane Reset Potential, $v_{reset}$ & -60mV \\
Refractory Period, $\delta_{ref}$ & 5ms \\
Adaptive Threshold Time Constant, $\tau_{theta}$ & $10^{7}$ms\\
Adaptive Threshold Voltage Increment, $\theta_{+}$ & 0.05 \\
Post-Synaptic Learning Rate, $\eta_{post}$ & $10^{-2}$ (MNIST)\\
& $4\times10^{-3}$ (F-MNIST)\\
Pre-Synaptic Learning Rate, $\eta_{pre}$ & $10^{-4}$ (MNIST)\\
&$4\times10^{-5}$ (F-MNIST)\\
Normalization Factor, $w_{norm}$ & 78.4 \\
No. of Excitatory Neurons, $n_{neurons}$& 225 / 400 (MNIST)\\
&400 (F-MNIST) \\
Static Inhibitory Synaptic Weight, $w_{recurrent}$ & -120 (MNIST)\\
& -250 (F-MNIST) \\
A-STDP Weight-Percentile Hyperparameter, $\alpha$ & 98 \\
A-STDP Non-Linearity Hyperparameter, $\sigma$ & 2 \\
\hline \hline
\end{tabular}\\ 
\end{table}

\begin{figure}[htp]
\centering
\subfigure[]{\includegraphics[width=0.35\textwidth]{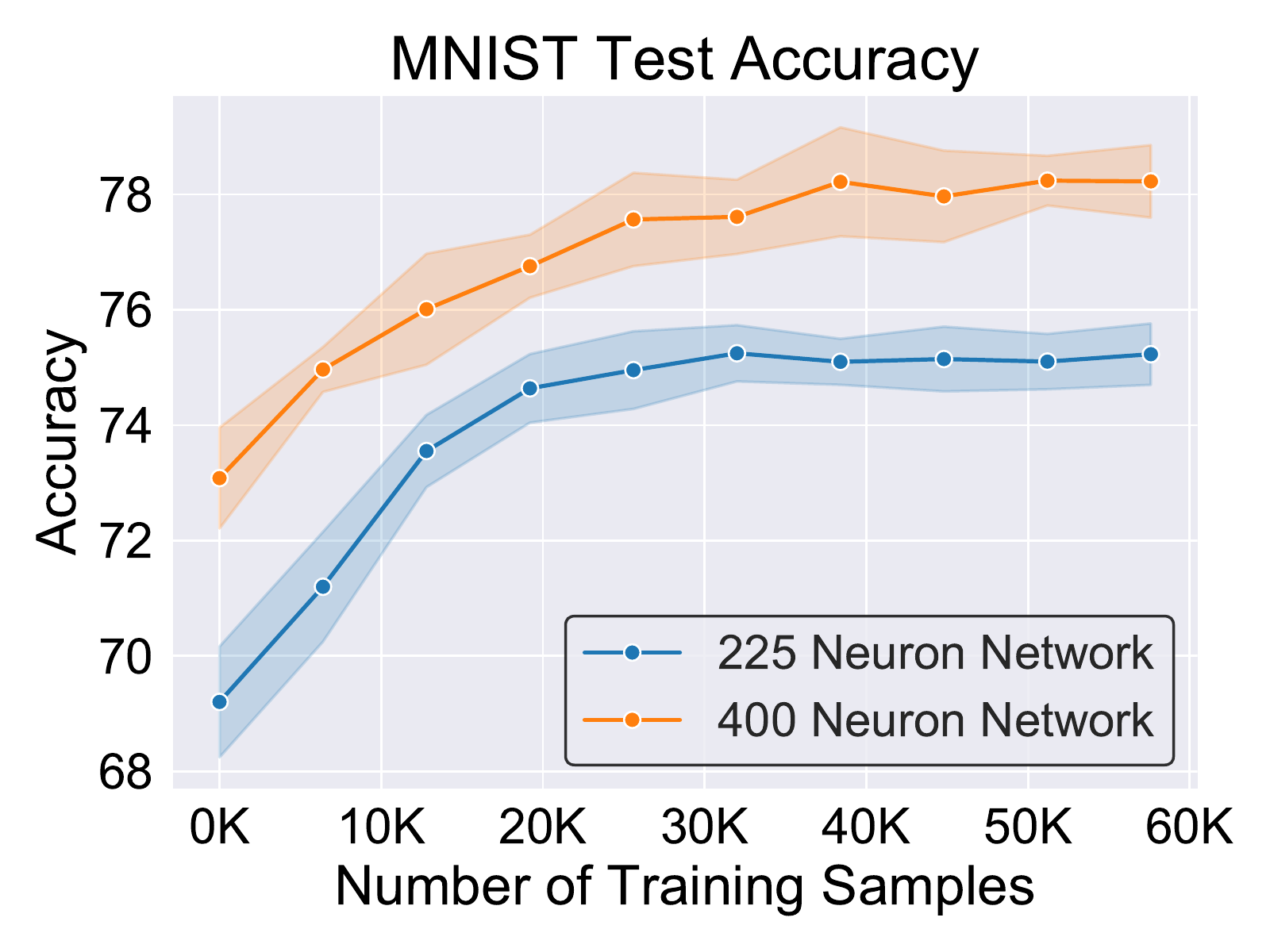}}
\subfigure[]{\includegraphics[width=0.35\textwidth]{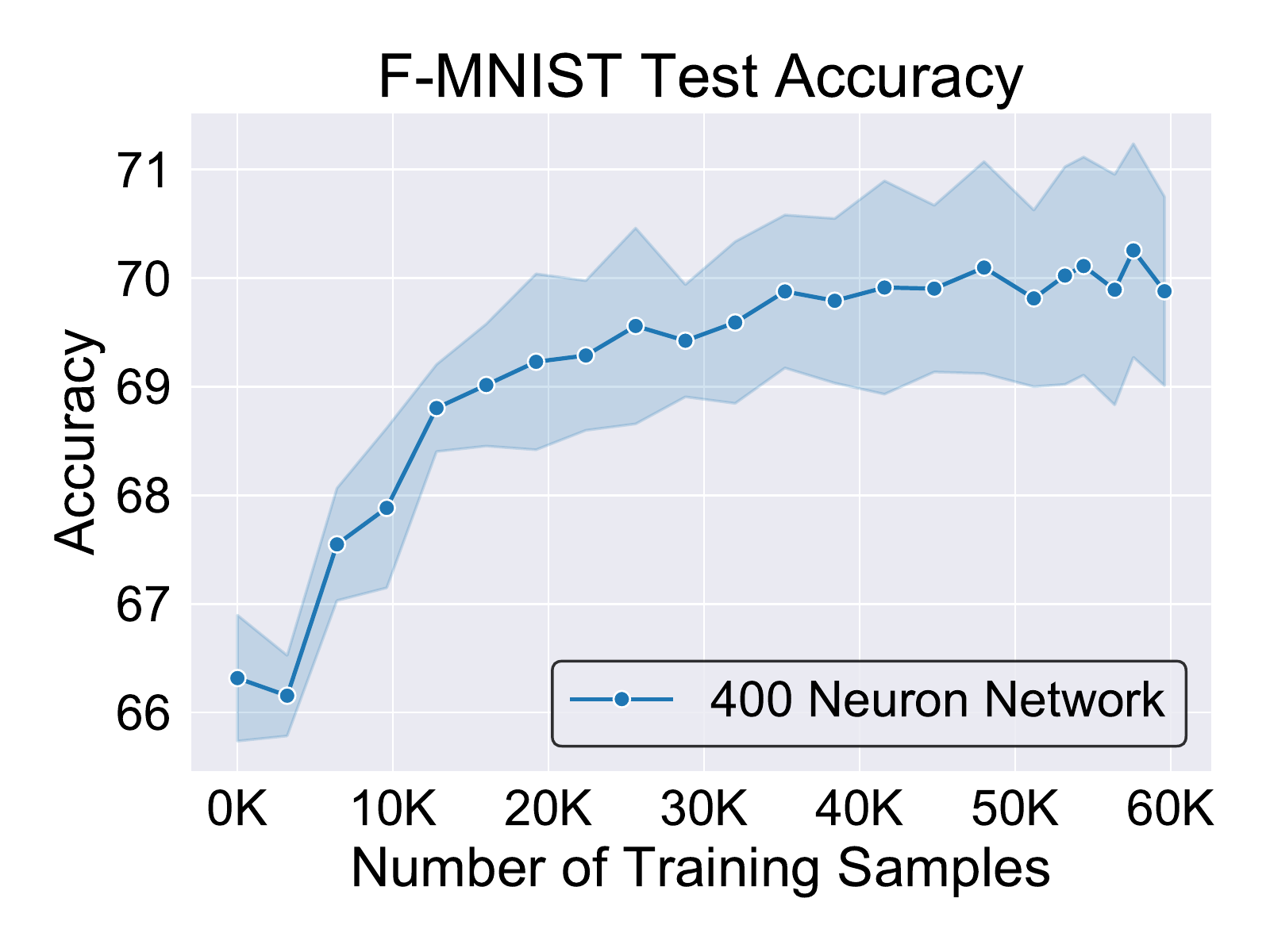}}

\caption{Improvement of test accuracy during re-learning is depicted as a function of the training samples using A-STDP learning rule on the (a) MNIST (225 and 400 neuron networks) and (b) F-MNIST datasets (400 neuron network). Mean and standard deviation of the accuracy is plotted for 80\% fault simulation in the networks.}
\label{fig:accuracy_astdp}
\end{figure}

The network is first trained with sole STDP learning rule for 2 epochs and the maximum test accuracy network is chosen as the baseline model. Subsequently, faults are introduced by randomly deleting synapses (from the Input to the Output Layer) post-training. Each synaptic connection was assigned a deletion probability, $p_{del}$, to decide whether the connection would be retained in the faulty network. For this work, $p_{del}$ was varied between 0.5 - 0.9 to analyze the network and re-train after introducing faults. Note that A-STDP learning rule is only used during this self-repair phase. It is worth mentioning here, that weight normalization by factor $w_{norm}$ (mentioned in Section III-B) is used before starting the re-training process. This helps to adjust the relative magnitude of firing threshold relative to the weights of the neurons (since the resultant magnitude diminishes due to fault injection). 

\begin{figure}[!t]
\centering
MNIST \linebreak
\medskip
\subfigure[Baseline Network]{\includegraphics[trim={2.5cm 0 0 0}, clip,  width=0.24\textwidth]{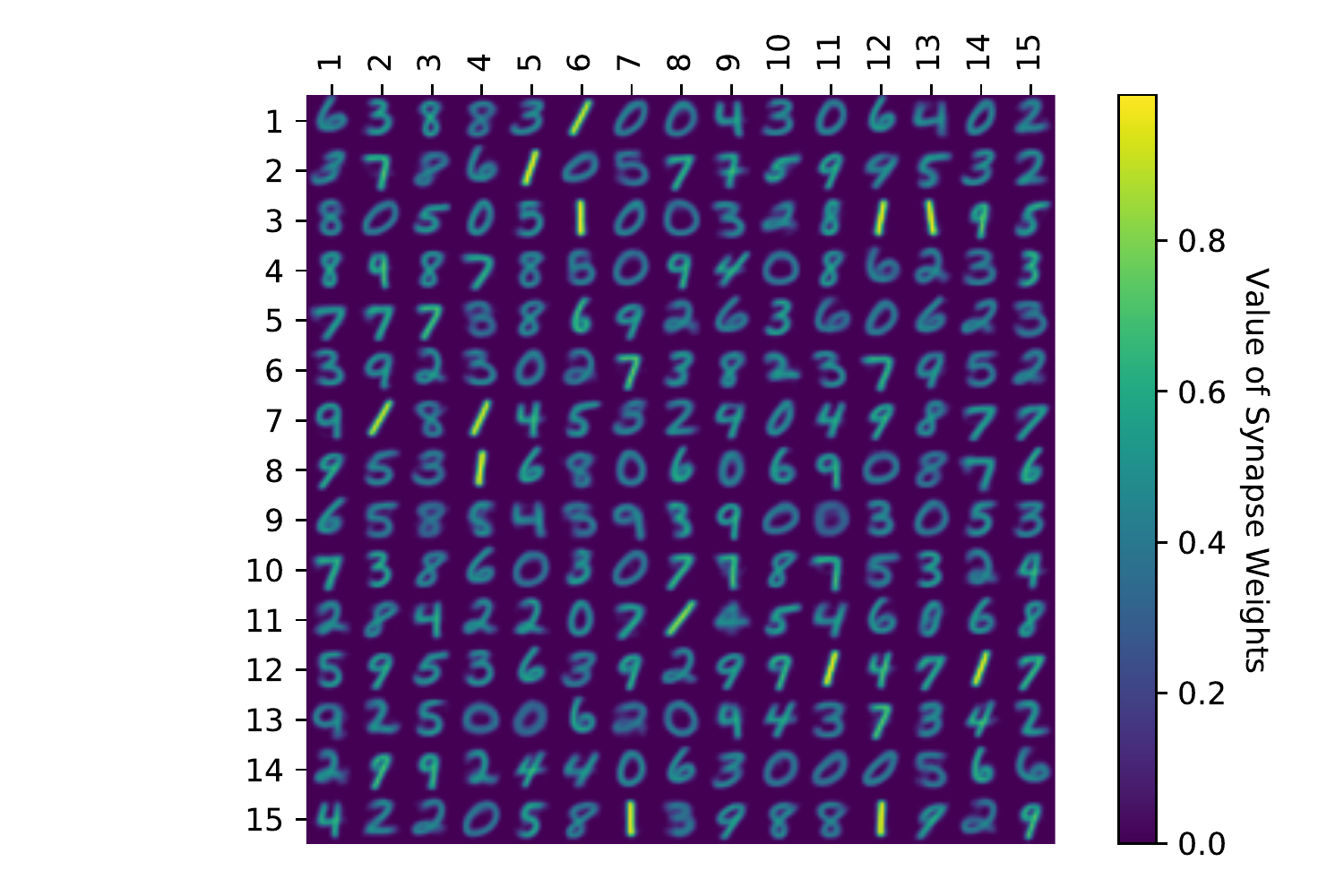}}
\hfil
\subfigure[After A-STDP Re-learning]{\includegraphics[trim={2.5cm 0 0 0}, clip, width=0.24\textwidth]{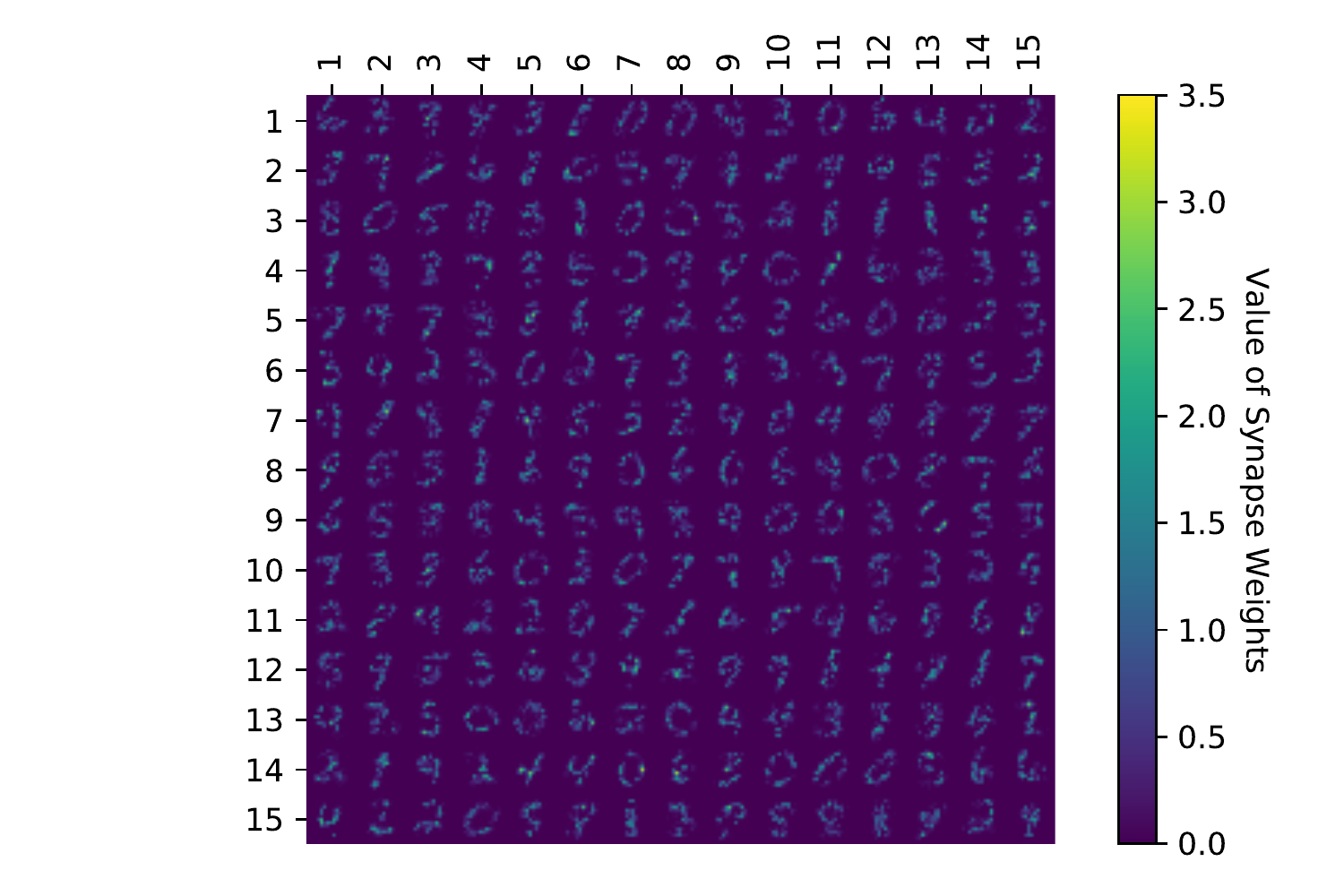}}
F-MNIST \linebreak
\subfigure[Baseline Network]{\includegraphics[trim={2.5cm 0 0 0}, clip, width=0.24\textwidth]{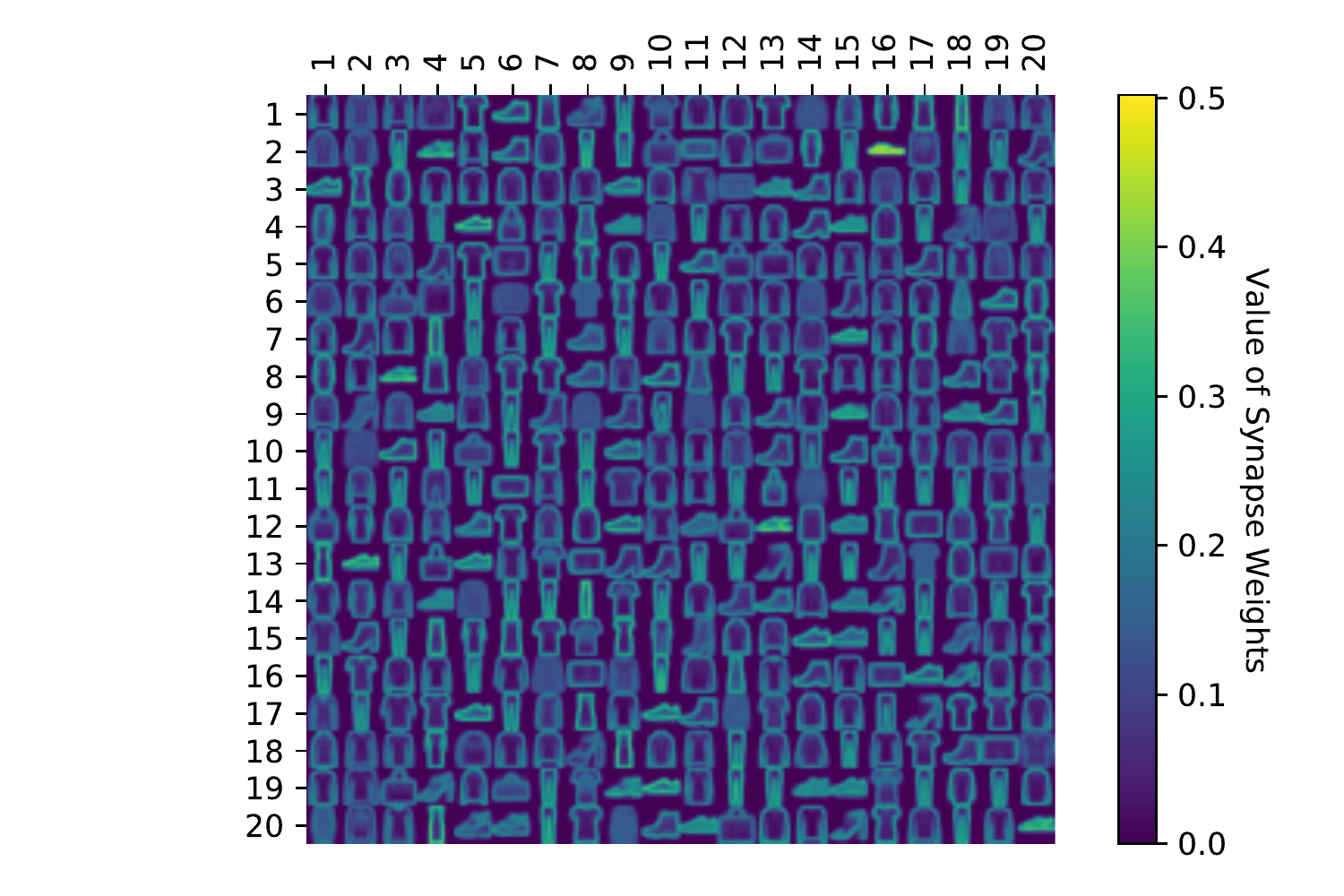}}
\subfigure[After A-STDP Re-learning]{\includegraphics[trim={2.5cm 0 0 0}, clip, width=0.24\textwidth]{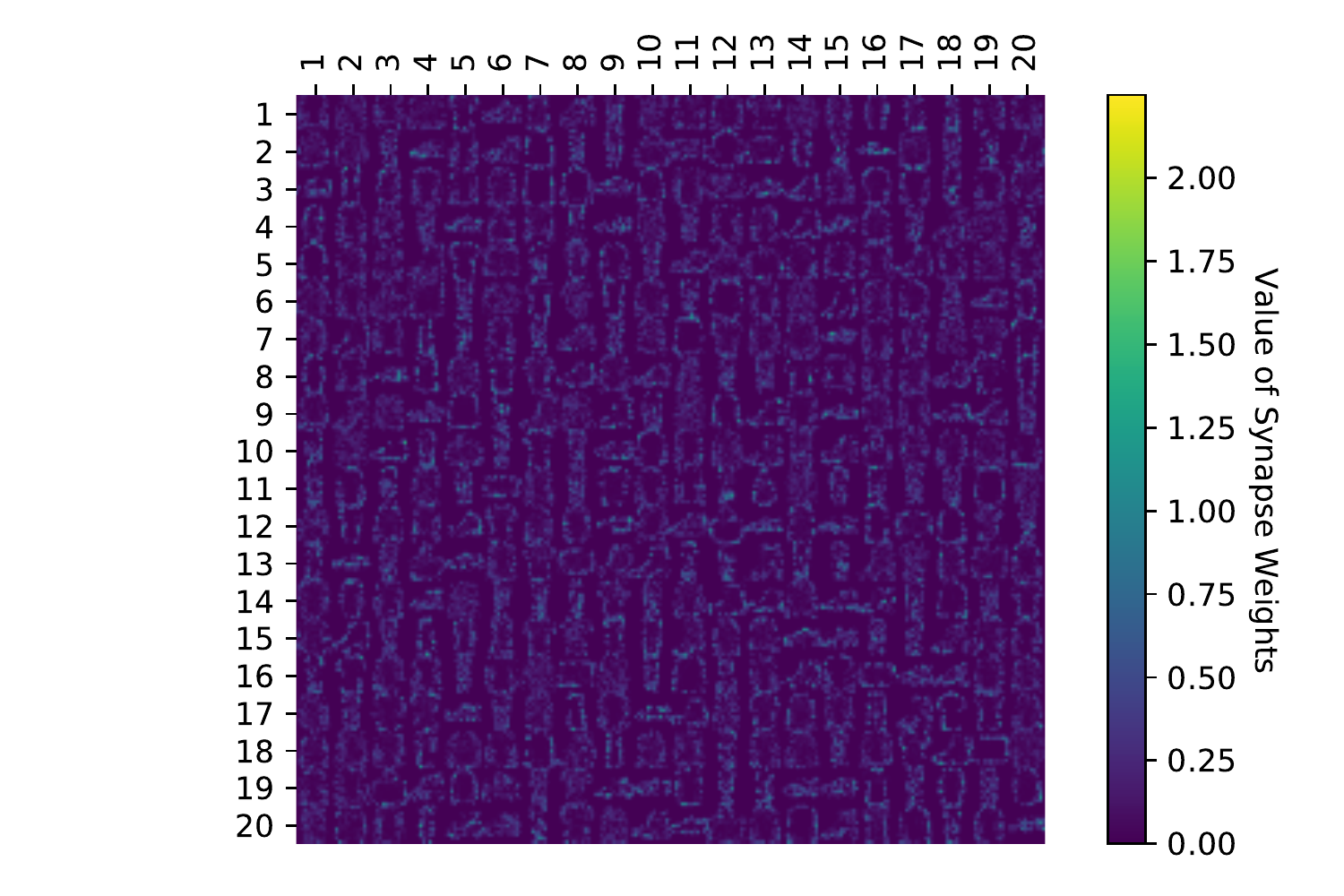}}
\caption{(a-d) Initial and self-repaired weight maps of the 225 (400) neuron network trained on MNIST (F-MNIST) dataset corresponding to 80\% fault simulations.}
\label{fig:Weight_map_relearn_1}
\end{figure}

\begin{table*}[!t]
\renewcommand{\arraystretch}{1}
\caption{Self-Repair Results for A-STDP Enabled SNNs}
\label{table:relearn_wstdp_table_results}
\centering
\begin{tabular}{  c c c c | c  | c c  } 
\hline
\hline
\\
 \begin{minipage}[t]{0.1\textwidth}\center{\textbf{Network Description}}\end{minipage} & 
 \begin{minipage}[t]{0.075\textwidth}\center{\textbf{Fault \\Probability}} \end{minipage} & 
  \begin{minipage}[t]{0.126\textwidth}\center{\textbf{Accuracy after fault \\ injection (\%)}}\end{minipage} &
 \begin{minipage}[t]{0.126\textwidth}\center{\textbf{Accuracy after weight \\normalization (\%)}}\end{minipage} &
 \begin{minipage}[t]{0.126\textwidth}\center{\textbf{Accuracy after \\STDP \\re-training (\%)}}\end{minipage} & 
 \begin{minipage}[t]{0.126\textwidth}\center{\textbf{Accuracy after \\A-STDP \\re-training (\%)}}\end{minipage} & 
 \begin{minipage}[t]{0.075\textwidth}\center{\textbf{Accuracy Gain from A-STDP}} \end{minipage}
 \\
 \\
\hline
\multicolumn{7}{c}{\textbf{MNIST Dataset}}  \\ 
\hline
\\
\multirow{5}{*}{\begin{minipage}[t]{0.18\textwidth}225 Excitatory Neurons \\ Baseline Accuracy = 89.53\%\end{minipage}} & 
50\%  & $75.41 \pm 2.28$ & $83.04 \pm 0.49 $ & $ 76.69 \pm 0.98 $& $84.06 \pm 0.70$  & 1.02  \\
&60\% & $70.72 \pm 1.41$ & $80.25 \pm 0.70$ & $73.85 \pm 1.18 $ & $82.19 \pm 0.28$ & 1.95 \\
& 70\% & $61.86 \pm 2.03$ &$76.06 \pm 1.22$& $70.57 \pm 0.48 $ &  $79.39 \pm 1.13$  & 3.33 \\
& 80\% & $30.42 \pm 3.26$ & $69.13 \pm 0.85$ & $ 67.42 \pm 1.37 $ &  $75.42 \pm 0.48$ & 6.29 \\
& 90\% & $9.92 \pm 0.11$ & $56.69 \pm 1.45$ &$ 61.40 \pm 1.38 $ & $65.57 \pm 1.28$   & 8.89 \\ 
\\
\hline
\\
\multirow{5}{*}{\begin{minipage}[t]{0.18\textwidth}400 Excitatory Neurons \\ Baseline Accuracy = 92.02\%\end{minipage}} 
& 50\% & $79.32 \pm 2.57$ & $85.56 \pm 0.24$ & $ 80.96\pm1.24 $ &  $87.16 \pm 0.12$ & 1.59 \\
& 60\% & $73.01 \pm 2.15$ &$82.61 \pm 0.22$ & $ 79.12 \pm 1.28 $ &   $85.17 \pm 0.33$ & 2.56 \\
& 70\% & $61.20 \pm 1.10$ & $79.77 \pm 0.61$ & $ 77.51 \pm 0.62 $ &  $83.00 \pm 0.40$  & 3.22\\
& 80\% & $30.18 \pm 2.60$ & $73.08 \pm 0.87$  &$ 73.26 \pm 1.16 $ &  $78.68 \pm 0.73$ & 5.58 \\
& 90\% & $9.80 \pm 0.27$ & $59.90 \pm 1.16$& $ 67.80 \pm 0.77 $ &$68.85 \pm 0.48 $  &8.95 \\ 
\\
\hline
\multicolumn{7}{c}{\textbf{Fashion-MNIST Dataset}} \\ 
\hline
\\
\multirow{5}{*}{\begin{minipage}[t]{0.18\textwidth}400 Excitatory Neurons \\ Baseline Accuracy = 77.35\%\end{minipage}}  &   
50\% & $58.62 \pm 1.24$ &  $73.85 \pm 0.50$ &  $73.51 \pm 0.30 $ &  $75.88 \pm 0.38$ & 2.02 \\
& 60\% & $39.12 \pm 1.19$ &  $71.85 \pm 1.36$ &  $72.23 \pm 0.60 $ &  $75.16 \pm 0.49$ & 3.31 \\
& 70\% & $16.61 \pm 0.69$ &  $70.21 \pm 0.44$ &  $70.63 \pm 0.70 $ &  $73.14 \pm 0.44$ & 2.93 \\
& 80\% & $10.00 \pm 0.22$ & $66.32 \pm 0.58$ &  $68.80 \pm 0.47 $ &  $70.82 \pm 0.57$ & 4.51 \\
& 90\% & $10.04 \pm 0.28$ &  $60.24 \pm 0.86$ &  $63.92 \pm 0.77 $ &  $65.49 \pm 0.40$ & 5.25 \\
\\
\hline
\hline
\end{tabular}
\end{table*}

Fig. \ref{fig:accuracy_astdp} shows the test classification accuracy as a function of re-learning epochs for a 225 / 400 neuron network with $80\%$ probability for faulty synapses. After the faults are introduced, the network accuracy improves over time during the self-repair process. The mean and standard deviation of test accuracy from 5 independent runs are plotted in Fig. \ref{fig:accuracy_astdp}. Fig. \ref{fig:Weight_map_relearn_1} depicts the initial and self-repaired weight maps of the 225 (MNIST) and 400 (F-MNIST) neuron networks, substantiating that original learnt representations are preserved during the re-learning process. Table \ref{table:relearn_wstdp_table_results} summarizes our results for all networks with varying degrees of faults. The numbers in parentheses denote the standard deviation in accuracy from the 5 independent runs. Since sole STDP learning resulted in accuracy degradation for most of the runs, the accuracy is reported after 1 re-learning epoch. For some cases, some accuracy improvement through normal STDP was also observed. The maximum accuracy is reported for the A-STDP re-training process. After repair through A-STDP, the network is able to achieve accuracy improvement across all level of faults, ranging from 50\% - 90\%. Interestingly, A-STDP is able to repair faults even in a 90\% faulty network and improve the testing accuracy by almost 9\% (5\%) for the MNIST (F-MNIST) dataset. Further, the accuracy improvement due to A-STDP scales up with increasing degree of faults. Note that the standard deviation of the final accuracy over 5 independent runs is much smaller for A-STDP than normal STDP re-training, signifying that the astrocyte enabled self-repair is consistently stable, irrespective of the initial fault locations. 




\section{Discussion}
The work provides proof-of-concept results toward the development of a new generation of neuromorphic computing platforms that are able to autonomously self-repair faulty non-ideal hardware operation. Extending beyond just unsupervised STDP learning, augmenting astrocyte feedback in supervised gradient descent based training of SNNs needs to be explored along with their implementation on neuromorphic datasets \cite{nmnist}. In this work, we also focused on aspects of astrocyte operation that would be relevant from a macro-modelling perspective for self-repair. Further investigations on understanding the role of neuroglia in neuromorphic computing can potentially forge new directions related to synaptic learning, temporal binding, among others. 

\section*{Data Availability Statement}

The original contributions presented in the study are included in the article, further inquiries can be directed to the corresponding author/s.

\section*{Author Contributions}

AS developed the main concepts. MR, SL and NI performed all the simulations. All authors assisted in the writing of the paper and developing the concepts.

\section*{Acknowledgments}

The work was supported in part by the National Science Foundation grants BCS \#2031632, ECCS \#2028213 and CCF \#1955815.





%






\end{document}